%% file: acl_latex.tex
\documentclass[11pt]{article}

\usepackage[preprint]{acl}

\usepackage{times}
\usepackage{latexsym}

\usepackage[T1]{fontenc}

\usepackage[utf8]{inputenc}

\usepackage{microtype}

\usepackage{inconsolata}

\usepackage{graphicx}

\usepackage{xspace}
\usepackage{multirow}
\usepackage{makecell}
\usepackage{booktabs}
\usepackage{pifont}
\usepackage{amssymb}
\usepackage{tcolorbox}
\usepackage[dvipsnames]{xcolor}
\usepackage{xcolor}
\newcommand{\cmark}{\color{green}\checkmark}       
\newcommand{\xmark}{\color{red}\ding{55}} 
\newcommand{\benchmark}{{WiserUI-Bench}\xspace}

\usepackage[table]{xcolor}
\usepackage{array}
\definecolor{overallbg}{RGB}{244,244,244}
\definecolor{besttype}{RGB}{219,238,252} 
\definecolor{secondtype}{RGB}{255,242,204}
\definecolor{newg2}{RGB}{52,127,89}
\definecolor{newb2}{RGB}{95,176,254}

\title{Do MLLMs Capture How Interfaces Guide User Behavior? \\ A Benchmark for Multimodal UI/UX Design Understanding}

\newcommand*\samethanks[1][\value{footnote}]{\footnotemark[#1]}

\author{
 \textbf{Jaehyun Jeon\textsuperscript{1}\thanks{Currently at NC AI}},
 \textbf{Min Soo Kim\textsuperscript{1}},
 \textbf{Jang Han Yoon\textsuperscript{1}},
 \textbf{Sumin Shim\textsuperscript{1}},
\\
 \textbf{Yejin Choi\textsuperscript{1}},
 \textbf{Hanbin Kim\textsuperscript{1}},
 \textbf{Dae Hyun Kim\textsuperscript{1}\thanks{Co-corresponding authors}},
 \textbf{Youngjae Yu\textsuperscript{2}\samethanks}
\\
 \textsuperscript{1}Yonsei University \quad
 \textsuperscript{2}Seoul National University
\\
 \texttt{\{jaehyun.jeon, dhkim16\}@yonsei.ac.kr}\quad
\texttt{youngjaeyu@snu.ac.kr}
}

\begin{document}
\maketitle
\begin{abstract}
User interface (UI) design goes beyond visuals to shape user experience (UX), underscoring the shift toward UI/UX as a unified concept.
While recent studies have explored UI evaluation using Multimodal Large Language Models (MLLMs), they largely focus on surface-level features, overlooking how design choices influence user behavior at scale.
To fill this gap, we introduce \benchmark
, a novel benchmark for \textit{multimodal understanding of how UI/UX design affects user behavior}, built on 300 real-world UI image pairs from industry A/B tests, with empirically validated winners that induced more user actions. For future design progress in practice, post-hoc understanding of why such winners succeed with mass users is also required; we support this via expert-curated key interpretations for each instance.
Experiments across multiple MLLMs on \benchmark for two main tasks, (1) predicting the more effective UI image between an A/B-tested pair, and (2) explaining it post-hoc in alignment with expert interpretations, show that models exhibit limited understanding of the behavioral impact of UI/UX design.
We believe our work will foster research on leveraging MLLMs for visual design in user behavior contexts.
\end{abstract}


\section{Introduction}

User interface (UI) design is central to application development. While visual aesthetics lay its foundation, the fundamental goal is to steer user behavior during interaction, such as encouraging sign-ups or purchases, and thereby boost service revenue \citep{fogg2002persuasive}. This behavioral focus expands to \textit{UI/UX design}, uniting visual form with user experience (UX) considerations \citep{norman1988psychology}.

\begin{figure}[t]
  \centering
  \includegraphics[width=0.98\columnwidth]{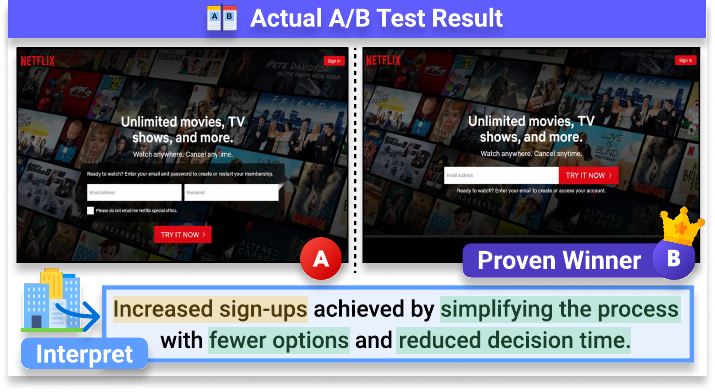}
  \vspace{-0.5em}
  \caption{A real-world, behavior-aware design decision example from \benchmark, grounded in A/B test results, illustrating how UI changes steer user actions.}
  \label{fig:motivation}
\end{figure}


Practitioners validate such decisions through large-scale A/B tests that randomly assign users to alternative UI variants and measure which design more effectively drives the desired action \citep{kohavi2009controlled}. 
For example, as shown in Figure \ref{fig:motivation}, version B achieves higher sign-ups overall, interpreted as reduced user friction from fewer options. Such interpretation reflects an abductive reasoning process \citep{hobbs1993interpretation, bhagavatula2020abductivecommonsensereasoning}, inferring the most plausible explanations for aggregate outcomes that are used to guide subsequent design decisions \citep{quin2024b}.
Given that predicting and explaining these results is nontrivial even for humans, hinging on complex cognitive processes underlying user experience, assessing \textit{multimodal understanding of UI/UX design} becomes a crucial challenge, requiring not only identifying visual elements but also reasoning about how they influence user behavior.



\begin{figure*}[t]
\centering
\includegraphics[width=0.98\textwidth]{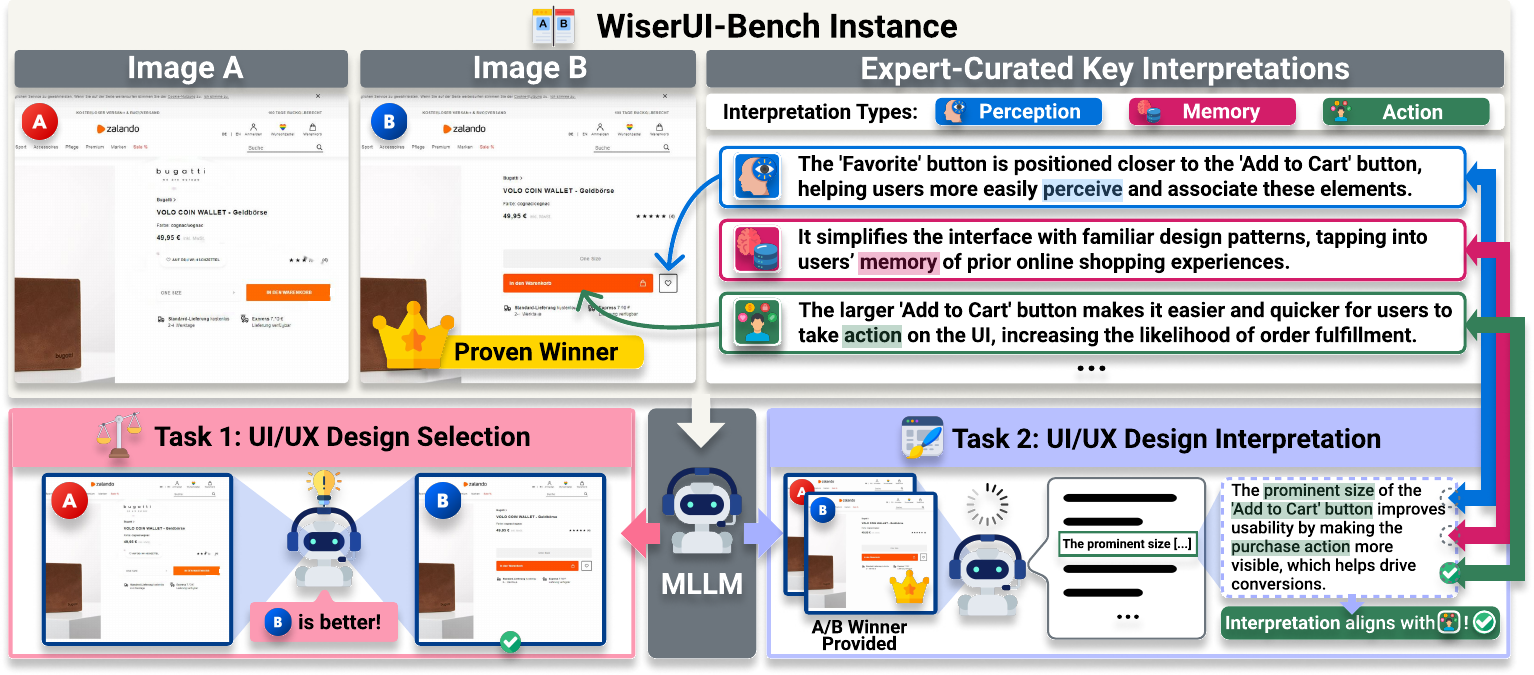} 
\vspace{-0.5em}
\caption{Overview of \benchmark and two main tasks. Each instance contains a UI image pair with verified A/B test winner and expert-curated key interpretations explaining it. These span the three cognitive UX dimensions; the example includes all three, though most cover fewer. MLLMs are evaluated on (1) selecting the more effective UI/UX design by predicting the verified winner, and (2) explaining the effectiveness of a given winner, measured by whether the model captures each expert interpretation.}
\label{fig:benchmark_task}
\end{figure*}

Recent efforts have explored how to evaluate UI design quality with Multimodal Large Language Models (MLLMs); however, they all fall short in capturing the behavioral impact of UI/UX design, as summarized in Table \ref{tab:benchmark_comparison}. While some focus solely on basic visual, heuristic-based attributes \citep{yang2024uisgpt, wu2024uiclip}, others rely on handwritten critiques of individual UIs, lacking validation against real user data and offering limited coverage of cognitive aspects \citep{duan2024uicrit}.

To address these limitations, we introduce \textbf{\benchmark} (Pairwise UI/UX Design Understanding Benchmark), a novel benchmark for evaluating a model's multimodal understanding of how UI/UX design affects user behavior. \benchmark comprises 300 real-world UI image pairs with winning variants verified by A/B tests from actual companies, spanning varied contexts and sourced from reliable platforms. Each pair includes expert-curated key interpretations, 684 in total, written post-hoc with knowledge of the winner to explain its effectiveness, covering core behavioral drivers. An example instance is illustrated in the upper part of Figure \ref{fig:benchmark_task}. Grounded in large-scale real user behavior data, our benchmark captures authentic usage and preference patterns, providing a reliable basis for evaluation. Each interpretation is tagged with one of the three cognitive UX dimensions articulated by \citet{norman2007emotional}: perception (what users notice), memory (what users remember), and action (how users act).




Our benchmark supports two core tasks, illustrated in the lower part of Figure \ref{fig:benchmark_task}: (1) selecting the more effective UI/UX design from a given pair by predicting the A/B test-verified winner, and (2) interpreting the UI/UX effectiveness of the given winner by assessing alignment between model-generated and expert-curated interpretations. This task setup captures the essential capabilities for UI/UX assessment, not only to identify which design better guides user behavior, but also to reliably explain why in a post-hoc manner.

Experiments on a broad range of proprietary and open-source MLLMs on \benchmark show that existing models struggle with visual reasoning in user behavior contexts. Models perform near-random on selection, and their post-hoc interpretations remain insufficient to reach expert level. We expect that \benchmark and our novel UI/UX-focused tasks will spur further research on this critical and underexplored, user-centered area, advancing behavior-aware reasoning about visual design in MLLMs.


\begin{table*}[t]
\centering
\resizebox{\linewidth}{!}{
\begin{tabular}{c|ccccccc}
\toprule
Benchmark & \# Samples & Unit & Source Quality & Source Type & \makecell[c]{Large-Scale User \\ Validation} & \makecell[c]{Expert \\Interpretation} & Evaluation Objective \\ 
\midrule
\makecell[c]{\citet{yang2024uisgpt}} & 382 & Single & Legacy UI & Mobile & \xmark & \xmark & \makecell[c]{Guideline Violation\\ Detection} \\ 

\makecell[c]{BetterWeb \\ \citep{wu2024uiclip}} & 892 & Pair & Synthetic & Mobile & \xmark & \xmark & \makecell[c]{Basic Visual Quality} \\ 

\makecell[c]{UICrit \\ \citep{duan2024uicrit}} & 983 & Single & Legacy UI & Mobile & \xmark & \cmark & \makecell[c]{UI Design Critique} \\ 
\midrule
\makecell[c]{\textbf{\benchmark} \\ \textbf{(Ours)}} & 300 & Pair & \makecell[c]{Actual \\Production} & Mobile + Web & \makecell[c]{\cmark \\ (Real A/B Test)} & \cmark & \makecell[c]{UI/UX Design Understanding \\(User Behavior-Focused)} \\ 

\bottomrule
\end{tabular}}
\caption{Comparison between \benchmark and existing UI evaluation benchmarks.}
\label{tab:benchmark_comparison}
\end{table*}

In summary, our main contributions are:
\begin{itemize}
    \setlength{\topsep}{0.3pt}
    \setlength{\itemsep}{0.3pt}
    \item We introduce a novel and challenging task of multimodal UI/UX understanding that centers on the behavioral impact of design.
    \item We release \benchmark, a benchmark of 300 real-world UI pairs uniquely constructed from verified results of large-scale industry A/B tests, along with 684 carefully curated expert key interpretations.
    \item Through comprehensive experiments and analyses across diverse MLLMs, we uncover their limited ability to understand the behavioral impact of UI/UX design, highlighting a critical gap in current visual reasoning.
\end{itemize}

\section{Related Work}

\subsection{Visual Reasoning}

Visual reasoning benchmarks for Multimodal Large Language Models (MLLMs) evaluate models' ability to process visual information and reason at a cognitive level. While some target general tasks \citep{yue2023mmmu, liu2024mmbench, fu2024mmecomprehensiveevaluationbenchmark, li2024seed}, others focus on specific domains like mathematics or science \citep{hao2025mllmsreasonmultimodalityemma, lu2024mathvista, xu2025visulogicbenchmarkevaluatingvisual}. However, few benchmarks address reasoning about human behavior or preference patterns in visual contexts. Moreover, although some studies explore visual reasoning across multiple images \citep{cheng2025evaluatingmllmsmultimodalmultiimage, zhao2024benchmarkingmultiimageunderstandingvision}, such cases remain rare. Our task incorporates both aspects, reasoning over paired UI images and aligning with behavior patterns, as captured by actual A/B test outcomes.

\subsection{UI Evaluation}
Recent advances have driven progress in evaluating UI quality with MLLMs. \citet{yang2024uisgpt} propose a benchmark for detecting guideline violations in individual screens, while \citet{wu2024uiclip} generates synthetic UI pairs via perturbations such as color noise or spacing changes. However, both are limited to heuristic analysis of surface-level appearance. \citet{duan2024uicrit} evaluates single screens using expert-critique annotations, but it lacks validation with real-world user behavior and relies on RICO \citep{deka2017rico}, an outdated, mobile-only dataset that limits its relevance to today's production-level designs. Prior work also suggests that pairwise evaluation yields more reliable signals, particularly for subjective tasks like ours that target behavior patterns captured by real-world A/B tests \citep{pairwise, 10.1145/3479863}. In this context, \benchmark offers a stronger foundation by grounding design effectiveness in large-scale user behavior data and requiring models to engage in deeper, nuanced reasoning over both visuals and UX in a pairwise format.

\subsection{Simulation of Human Behavior}

Efforts to simulate human behavior with AI are ongoing. \citet{park2023generative} places LLM-driven characters in a sandbox town, producing socially plausible interactions. \citet{lu2025promptingneedevaluatingllm} moves closer to real behavior by using actual platform logs to predict next user actions from prior context, yet remains text-only. In the UI domain, \citet{lu2025uxagentsimulatingusabilitytesting} and \citet{wang2025agentabautomatedscalableweb} deploy persona-based agents to browse interfaces and examine how designs may steer user behavior; the former checks realism only via qualitative interviews, while the latter compares against an actual A/B test but only under a single scenario. By contrast, our work grounds evaluation in numerous real-user A/B test results and introduces diverse multimodal tasks that reason over UI variant image pairs to predict and explain user behavior.

\section{\benchmark} \label{main:benchmark}

We propose \benchmark, a new benchmark for evaluating multimodal understanding of UI/UX design, a key factor in crafting interfaces that influence user behavior.

\begin{figure*}[t]
  \centering
  \includegraphics[width=0.96\textwidth]{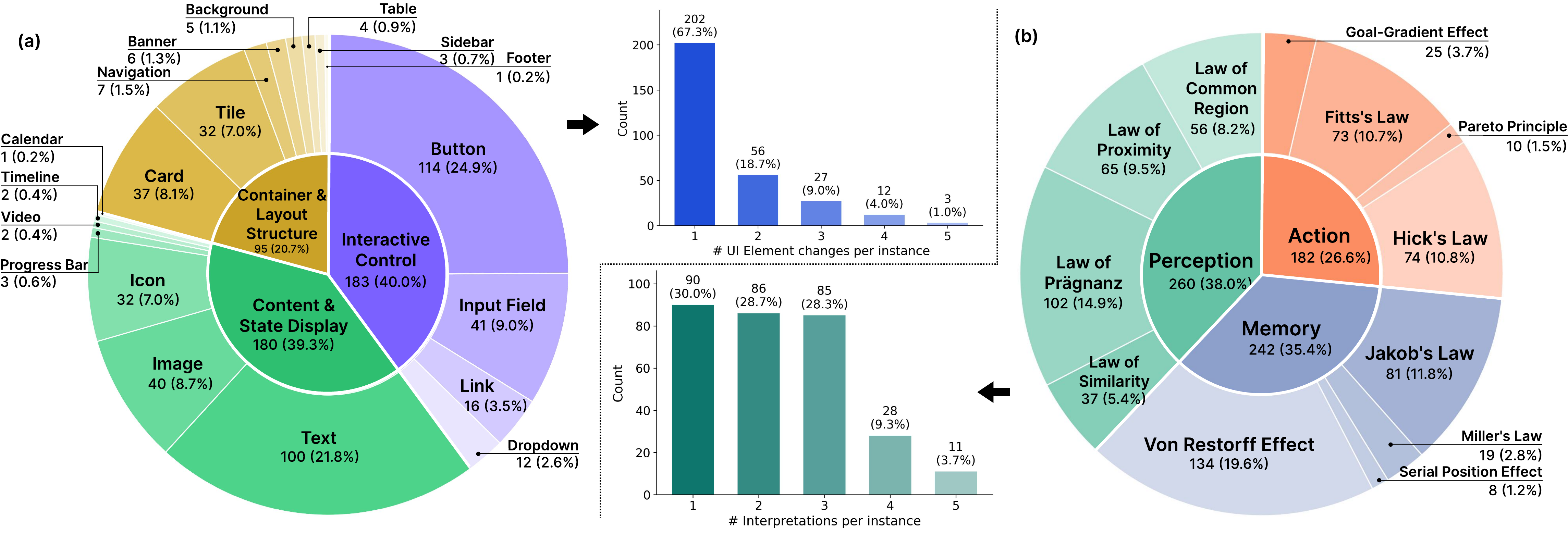}
  \vspace{-0.5em}
  \caption{Distribution of (a) UI change element types and (b) expert interpretations for \benchmark.}
  \label{fig:rationale_stat}
\end{figure*}

\begin{figure}
  \centering
  \includegraphics[width=0.98\columnwidth]{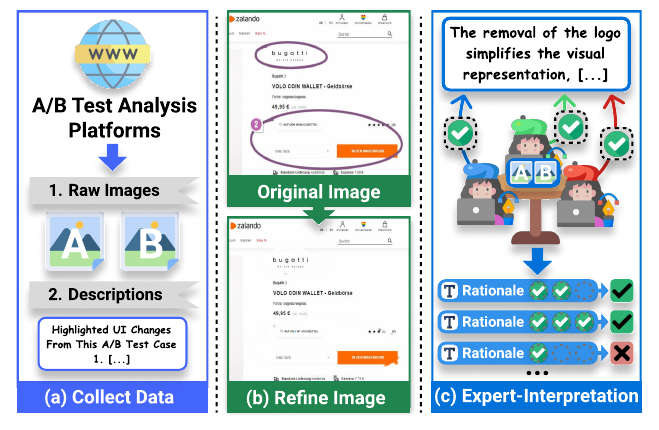}
  \vspace{-0.5em}
  \caption{Benchmark construction pipeline. (a) Raw A/B test data were collected from reliable platforms, (b) then UI images were refined for clarity. (c) Finally, key interpretations explaining the test outcomes were curated by UI/UX experts.}
  \label{fig:abtestbenchmark}
\end{figure}

\subsection{Benchmark Construction}

\paragraph{Data Sources}

To construct a benchmark that faithfully reflects real-world UI/UX design decisions, we collect UI images from industry-validated A/B testing data. Specifically, we aggregate data from widely recognized A/B testing and analysis platforms\footnote{\url{https://vwo.com/success-stories/}, \url{https://goodui.org/leaks/}, \url{https://abtest.design/}}, which showcase large-scale A/B tests conducted by companies across various industries and regions. These sources provide both UI variants and verified outcomes based on actual user behavior data. By grounding our benchmark in such real, rigorously tested cases, we ensure a credible basis for evaluating UI/UX design understanding in practical scenarios.


\paragraph{Data Curation Pipeline}
As illustrated in Figure \ref{fig:abtestbenchmark}, our curation process involves multiple stages to ensure high-quality annotations. First, we extract raw UI images and accompanying textual descriptions from our selected platform sources. Next, we refine these images by eliminating any visual markers that were externally added, such as arrows or circles, using inpainting techniques, to preserve a clean, unbiased evaluation of visual reasoning.

Finally, we construct key expert interpretations explaining the winning variant's superior performance. Such expert involvement is essential, as interpreting user behavior often requires deep domain expertise. Three experienced UI/UX experts aware of the A/B test outcomes independently annotated each instance, instructed to identify (i) the key UI modification and (ii) its corresponding behavioral implication. As a pre-defined quality control step, only interpretations where at least two experts independently converged on substantially overlapping findings were retained, yielding 684 final interpretations (562 with full, 122 with two-way agreement) from 845 initial candidates. Each interpretation was further tagged with the most relevant law from a curated set of 12 UX laws. Since user behavior is often shaped by multiple factors, a single instance may include multiple interpretations. See Appendix \ref{appx:annotation} for details.




\subsection{Benchmark Analysis}

Our pipeline produces a benchmark of 300 real-world UI image pairs, each with a definitive winner identified through A/B test results and annotated with expert-curated key interpretations of the core design factors driving user behavior, totaling 684 entries. We include detailed statistics in Appx. \ref{appx:stat}.

Ensuring contextual diversity is essential, as effective UI/UX strategies vary widely across use cases \citep{oinas2018persuasive}. For example, page types correspond to different stages of the user journey, requiring distinct strategies. To capture this, \benchmark spans 11 page types, 13 industry domains, and two device types.


Moreover, we annotate UI differences using element–attribute associations, specifying the UI element being modified and the corresponding attribute change (e.g., a \textit{button} with a \textit{position} change), to capture real-world UI variants with concurrent modifications. On average, each instance contains 1.53 element-level and 1.99 attribute-level annotations, spanning 19 element types and 14 attribute types. We further group the element types by their primary functional roles (Figure \ref{fig:rationale_stat}(a)): \textit{Interactive Control}, which directly elicits intentional user actions; \textit{Content \& State Display}, which primarily conveys information, semantics, or system state; and \textit{Container \& Layout Structure}, which organizes spatial layout and guides attention at a structural level.



\benchmark also exhibits diversity in the interpretations used to explain user behavior, as shown in Figure \ref{fig:rationale_stat}(b), with an average of 2.28 interpretations per instance. The tagged UX laws are grouped into three cognitive dimensions of UX, perception, memory, and action, following \citet{norman2007emotional}, with four laws assigned to each.

\section{Task Description}

Proper evaluation of a model's multimodal understanding of UI/UX design requires assessing not only its ability to predict which design more effectively guides user behavior, but also whether it can explain \textit{why} in a post-hoc manner, grounding its reasoning in visual and behavioral cues. To this end, we define two main tasks, selection and interpretation, as illustrated in Figure \ref{fig:benchmark_task}, both essential for demonstrating genuine understanding. Success requires identifying key visual differences between two UI variants, as well as inferring the underlying behavioral goal from context (e.g., promoting sign-ups) and how the interface supports that goal.

\paragraph{Task 1: UI/UX Design Selection}

The first task assesses whether the model can identify which of two UI images is more effective at influencing user behavior, with the correct answer determined by large-scale A/B testing run by real companies. Although the task is framed as a binary choice, solving it effectively requires nuanced visual reasoning beyond surface-level comparison.

\paragraph{Task 2: UI/UX Design Interpretation}

The second task evaluates a model's post-hoc interpretation ability on the observed results. Unlike the first task, it presents the winning UI/UX design upfront and asks the model to explain the factors contributing to its effectiveness in guiding user behavior. The model’s explanation is then compared to expert-curated key interpretations, with the degree of alignment serving as a measure of the quality of its expert-level interpretive understanding.

\section{Experiments} \label{main:experiment}

\paragraph{Models}

We evaluate a diverse set of MLLMs, spanning both \textit{proprietary} and \textit{open-source} models. On the proprietary side, we include o1 \citep{openai2024openaio1card}, GPT-5.1 \citep{openai2025gpt51}, GPT-4o \citep{openai2024gpt4ocard}, Claude 4.5 Sonnet \citep{anthropic2025claudesonnet45}, Claude 3.5 Sonnet \citep{anthropic2024claude3}, all noted for their strong visual reasoning. For open-source models, we evaluate widely adopted models such as Qwen3-VL (30B-A3B, 8B) \citep{bai2025qwen3}, Qwen2.5-VL (7B, 32B) \citep{bai2025qwen25vltechnicalreport}, InternVL-2.5 (8B, 38B) \citep{chen2025expandingperformanceboundariesopensource}, LLaVA-NeXT-7B \citep{liu2024improved} and LLaVA-OneVision-7B \citep{li2024llavaonevisioneasyvisualtask}. Refer to Appendix \ref{appx:setup} for details.

\subsection{UI/UX Design Selection} \label{sec:task1_result}



\subsubsection{Evaluation Metrics}

We report model accuracies across input-order variations of each UI image pair, reflecting the multi-image input setting. Specifically, we report: (1) \textit{First Accuracy (FA)}, when the more effective UI/UX design is presented first in the input; (2) \textit{Second Accuracy (SA)}, when it is presented second; and (3) \textit{Average Accuracy (AA)}, the mean of FA and SA, reflecting overall correctness. Finally, we compute (4) \textit{Consistent Accuracy (CA)}, which checks whether the model selects the same, correct UI across both input orders of each pair. This metric isolates the model’s core visual reasoning ability by ensuring that its choices are based on content rather than position bias, a well-known issue in pairwise comparison settings \citep{wang2025eliminatingpositionbiaslanguage, zhang2023gpt4visiongeneralistevaluatorvisionlanguage}. (Random guessing yields a CA of 25\%: since there's a 50\% chance of being correct in each order.) All evaluations are conducted over three independent runs, and we report the averaged results. Experiment details, including the prompts, are provided in Appendix \ref{appx:selection}.

\subsubsection{Results}

\begin{table*}[t]
\centering
\small
\resizebox{\linewidth}{!}{
\begin{tabular}{l|
>{\columncolor{overallbg}}c
>{\columncolor{overallbg}}c|
>{\columncolor{overallbg}}c
>{\columncolor{overallbg}}c|
cc|cc|cc|cc}
\toprule
& \multicolumn{4}{c|}{{\cellcolor{overallbg}}Overall}
& \multicolumn{2}{c|}{IC}
& \multicolumn{2}{c|}{CSD}
& \multicolumn{2}{c|}{CLS}
& \multicolumn{2}{c}{Mixed} \\
\# Cases
& \multicolumn{4}{c|}{{\cellcolor{overallbg}}(300)}
& \multicolumn{2}{c|}{(100)}
& \multicolumn{2}{c|}{(79)}  
& \multicolumn{2}{c|}{(49)}  
& \multicolumn{2}{c}{(72)}   \\
\midrule

& FA & SA & AA & CA
& AA & CA
& AA & CA
& AA & CA
& AA & CA \\
\midrule
Random
& 50.00 & 50.00 & 50.00 & 25.00
& 50.00 & 25.00
& 50.00 & 25.00
& 50.00 & 25.00
& 50.00 & 25.00  \\
\specialrule{0.6pt}{0.2ex}{0.3ex}
\specialrule{0.6pt}{0pt}{0.6ex}
o1
& 16.56 & 97.78 & 57.17 & 15.56
& 55.67 & \cellcolor{secondtype}16.00
& 55.70 & 11.81
& 56.46 & 11.56
& \underline{61.34} & \cellcolor{besttype}21.76 \\
GPT-5.1
& 35.67 & 81.33 & \underline{58.50} & \underline{33.33}
& \underline{58.67} & \underline{33.33}
& 59.28 & \cellcolor{secondtype}35.02
& 54.42 & 25.17
& 60.19 & \cellcolor{besttype}\underline{37.04} \\
GPT-4o
& 31.89 & 88.33 & \textbf{60.11} & 30.11
& \textbf{59.67} & \cellcolor{secondtype}31.00
& \textbf{61.60} & 29.96
& 51.70 & 17.69
& \textbf{64.81} & \cellcolor{besttype}\textbf{37.50} \\
Claude 4.5 Sonnet
& 34.33 & 79.33 & 56.83 & 32.33
& 53.50 & 29.00
& \underline{60.76} & \cellcolor{besttype}\textbf{37.97}
& 55.10 & \cellcolor{secondtype}\textbf{34.69}
& 58.33 & 29.17 \\
Claude 3.5 Sonnet
& 26.11 & 86.67 & 56.39 & 24.22
& 56.67 & 22.67
& 57.59 & \cellcolor{besttype}29.96
& 51.70 & 12.24
& 57.87 & \cellcolor{secondtype}28.24 \\
\midrule
Qwen3-VL-30B-A3B
& 16.67 & 88.00 & 52.33 & 15.00
& 53.33 & \cellcolor{secondtype}16.67
& 51.69 & 12.66
& 54.08 & \cellcolor{besttype}17.69
& 50.46 & 13.43 \\
Qwen3-VL-8B
& 19.67 & 87.67 & 53.67 & 19.33
& 51.67 & 16.00
& 57.38 & \cellcolor{besttype}23.63
& 49.66 & 16.33
& 55.09 & \cellcolor{secondtype}21.30 \\
Qwen2.5-VL-32B
& 33.67 & 81.44 & 57.56 & 31.00
& \underline{58.67} & \cellcolor{secondtype}31.67
& 56.54 & 28.69
& \textbf{58.84} & \underline{28.57}
& 56.25 & \cellcolor{besttype}34.26 \\
Qwen2.5-VL-7B
& 23.00 & 77.89 & 50.44 & 12.89
& 48.17 & 8.67
& 56.12 & \cellcolor{besttype}19.41
& 43.88 & 9.52
& 51.85 & \cellcolor{secondtype}13.89 \\
InternVL-2.5-38B
& 55.67 & 60.00 & 57.83 & \textbf{34.56}
& 56.17 & \textbf{34.33}
& 59.70 & \cellcolor{secondtype}\underline{36.29}
& 55.10 & \underline{28.57}
& 59.95 & \cellcolor{besttype}\underline{37.04} \\
InternVL-2.5-8B
& 58.33 & 45.67 & 52.00 & 21.33
& 52.17 & \cellcolor{secondtype}23.00
& 48.73 & 15.61
& \underline{57.14} & \cellcolor{besttype}25.85
& 51.85 & 22.22 \\
LLaVA-NeXT-7B
& 54.44 & 40.56 & 47.50 & 10.78
& 48.50 & \cellcolor{besttype}12.67
& 43.67 & 8.44
& 46.60 & 8.16
& 50.93 & \cellcolor{secondtype}12.50 \\
LLaVA-OneVision-7B
& 19.78 & 81.56 & 50.67 & 10.44 
& 50.00 & 11.00
& 54.43 & \cellcolor{besttype}13.50
& 47.28 & 3.40
& 49.77 & \cellcolor{secondtype}11.11 \\
\bottomrule
\end{tabular}}

\caption{Results of the UI/UX design selection task on \benchmark, shown overall and by UI change element types. Best and second-best model scores for each metric are marked in \textbf{bold} and \underline{underlined}, respectively. For CA, the best and second-best UI element type performances within each model are highlighted in \colorbox{besttype}{blue} and \colorbox{secondtype}{\rule{0pt}{0.5em}yellow}, respectively. All metrics are reported as percentages (\%).}
\label{tab:task1_result}
\end{table*}

\paragraph{Overall Performance}


Table~\ref{tab:task1_result} summarizes model performance on the UI/UX design selection task. Most AA scores hover only slightly above the 50\% random baseline for a binary choice. Moreover, they are inflated by strong position bias, especially in proprietary models and the Qwen-VL family, which tend to pick the second image. When we examine CA, the order-invariant metric, performance drops back to near-random levels. Interestingly, InternVL-2.5-38B scores slightly higher than proprietary models on CA, but still fall short of being adequate. Contrary to expectations, o1 shows the strongest bias and a lower CA than GPT-4o, suggesting that self-reasoning of o1 does not transfer well to this task. Most smaller (7B/8B) models perform significantly worse than larger ones, confirming the scaling trend. Overall, existing MLLMs struggle to consistently predict the more effective UI/UX design in real-world UI pairs.

\begin{table}[t]
\centering
\small  
\resizebox{\linewidth}{!}{
\begin{tabular}{l|l|cc|cc}
\toprule
Model & Method & FA & SA & AA & CA \\
\midrule
\multirow{6}{*}{\shortstack{GPT-4o}}
&Zero-Shot     & 31.89 & 88.33 & \underline{60.11} & 30.11 \\
& CoCoT         & 33.00 & 83.33 & 58.17 & 29.67 \\
& Self-Refine   & 32.44 & 82.67 & 57.56 & 29.89  \\
& DDCoT         & 36.56 & 86.89 & \textbf{61.72} & \underline{34.78}  \\
& MAD (R1)      & 51.22 & 67.44 & 59.33 & \textbf{39.00} \\
& MAD (R3)      & 34.44 & 80.89 & 57.67 & 29.89 \\
\midrule
\multirow{6}{*}{\shortstack{Claude 3.5 \\Sonnet}}
& Zero-Shot     & 26.11 & 86.67 & \underline{56.39} & 24.22 \\
& CoCoT         & 29.78 & 81.78 & 55.78 & \underline{27.22} \\
& Self-Refine   & 25.67 & 76.44 & 51.06 & 20.44 \\
& DDCoT         & 26.22 & 86.67 & \textbf{56.44} & 23.56 \\
& MAD (R1)      & 39.00 & 72.89 & 55.94 & \textbf{30.00} \\
& MAD (R3)      & 23.33 & 85.44 & 54.39 & 21.44 \\
\bottomrule
\end{tabular}}
\caption{Results of the UI/UX design selection task across reasoning strategies on \benchmark, figures in percentages (\%).}
\label{tab:task1_reason}
\end{table}

\paragraph{Performance Across UI Element Types}

The UI element-wise results in Table \ref{tab:task1_result} show performance variations across change type categories, while preserving overall trends: GPT-4o and InternVL-2.5-38B generally achieve the highest AA and CA, respectively. Focusing on CA, Mixed cases, where changes span multiple functional UI element groups, often yield relatively high performance, suggesting that aggregated multi-functional changes may provide richer behavioral cues than single-group modifications. Interactive Control (IC) and Content \& State Display (CSD) show broadly comparable performance overall, although their relative ranking varies across models. Across model families, GPT-based models show relative strength on Mixed cases, whereas Claude-based models do so on CSD, which often involves content-level changes (e.g., images or text). By contrast, Container \& Layout Structure (CLS) is the most challenging category, with sharply reduced CA across most models, reflecting limited sensitivity to structural layout changes, such as card or tile-level differences. Notably, Claude 4.5 Sonnet perform relatively better on CLS than other models.



\paragraph{Effect of Reasoning Strategies}

We additionally evaluate various reasoning strategies on GPT-4o and Claude 3.5 Sonnet to assess their effectiveness, including Contrastive Chain-of-Thought (CoCoT) \citep{zhang2024cocotcontrastivechainofthoughtprompting}, Self-Refine \citep{madaan2023self}, Duty-Distinct Chain-of-Thought (DDCoT) \citep{zheng2023ddcot}, and Multi-Agent Debate (MAD) \citep{liang-etal-2024-encouraging, du2023improving}. For MAD, we stop and moderate the debate after the first (R1) and third (R3) rounds. As shown in Table \ref{tab:task1_reason}, AA scores remain similar to or below the zero-shot setting, while CA scores often surpass it, especially for MAD (R1) across models, DDCoT on GPT-4o, and CoCoT on Claude 3.5 Sonnet. This gain is mainly due to reduced position bias, enabling more consistent decisions regardless of input order. MAD (R1) performs well by generating and moderating conflicting perspectives, which suits this human behavior pattern-driven task, benefiting from multi-perspective reasoning. However, excessive internal debate may hurt performance, as seen in the drop with MAD (R3).

\begin{figure*}[t]
  \centering
  \includegraphics[width=0.85\textwidth]{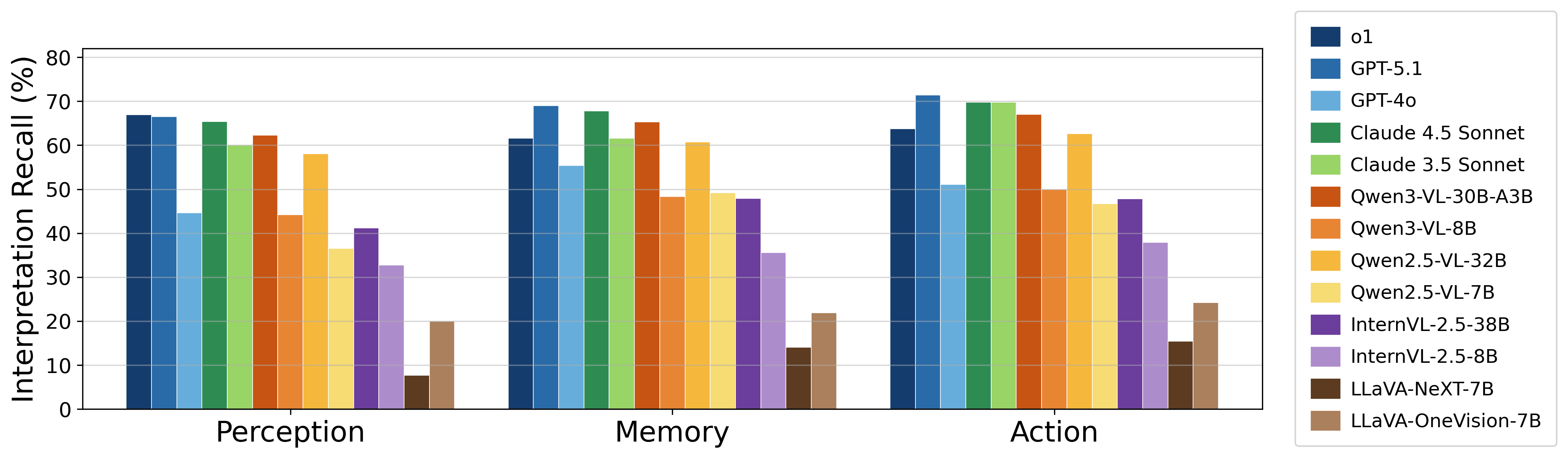}  
  \vspace{-0.5em}
  \caption{Results of the UI/UX design interpretation task by each UX dimension on \benchmark.}
  \label{fig:ux_rationale}
\end{figure*}

\begin{table}[t]
\centering
\small  
\resizebox{\linewidth}{!}{
\begin{tabular}{l|cc}
\toprule
\multirow{2}{*}{Model}
& Interpretation & Instance-level \\
& Recall & Recall \\
\midrule
o1 & 64.18 & 78.33 \\
GPT-5.1 & \textbf{68.71} & \underline{79.00} \\
GPT-4o & 50.15 & 66.67 \\
Claude 4.5 Sonnet & \underline{67.40} & \textbf{80.33} \\
Claude 3.5 Sonnet & 63.16 & 78.67 \\
\midrule
Qwen3-VL-30B-A3B & 64.62 & 74.00 \\
Qwen3-VL-8B & 47.22 & 56.00 \\
Qwen2.5-VL-32B & 60.23 & 73.00 \\
Qwen2.5-VL-7B & 43.71 & 61.67 \\
InternVL-2.5-38B
& 45.32 & 59.33 \\
InternVL-2.5-8B
& 35.09 & 51.67  \\
LLaVA-NeXT-7B
& 11.99  & 21.67  \\
LLaVA-OneVision-7B
& 21.78 & 33.33 \\

\bottomrule
\end{tabular}}
\caption{Results of the UI/UX design interpretation task on \benchmark, figures in percentages (\%).}
\label{tab:task2_result}
\end{table}

\subsection{UI/UX Design Interpretation} \label{sec:task2_result}

\subsubsection{Evaluation Metrics}

To judge how well a model’s generated interpretation explains the observed outcome, we compare it against expert-curated key interpretations for each instance. Since we let models produce free-form explanations, their outputs often differ substantially in wording and verbosity even when conveying similar meaning; accordingly, we prioritize semantic alignment over surface-level lexical overlap. Specifically, we instruct the model to enumerate all relevant factors in its explanation, and each expert interpretation is independently evaluated in a binary manner, whether it is fully captured by the explanation or not. We report two metrics: \textit{Interpretation Recall}, measuring the proportion of expert interpretations captured, and \textit{Instance‑level Recall}, which counts an instance as correct if at least one expert interpretation is captured. Binary inclusion judgments are made using a GPT-4o-based evaluator \citep{liu-etal-2023-g}, validated against human judgments on 1,000 random samples (83.0\% accuracy, Cohen's kappa 0.66). The prompts used, along with the results of BLEU, ROUGE, METEOR, and BERTScore \citep{papineni-etal-2002-bleu, lin-2004-rouge, banerjee-lavie-2005-meteor, zhang2020bertscoreevaluatingtextgeneration} are reported in the Appendix \ref{appx:interpret}.


\subsubsection{Results}

\paragraph{Overall Performance}



Table \ref{tab:task2_result} reports performance on the UI/UX design interpretation task. Recent models such as GPT-5.1 and Claude 4.5 Sonnet are among the strongest performers, although the overall results remain far from saturated. Some models, such as o1 and Claude 3.5 Sonnet, also show relatively strong post-hoc interpretation given the more effective variant, despite being weaker on selection in Task 1. By contrast, models such as GPT-4o and InternVL-2.5-38B, which were relatively strong on Task 1 selection, do not maintain the same level of advantage under this setting. Taken together, these observations suggest that selecting the better design and explaining its advantage are related but dissociable capabilities. Within the smaller-model regime, Qwen-VL variants align more closely with expert interpretations than other 7B/8B models, and larger Qwen-VL models consistently outperform their smaller counterparts. The persistent gap between Interpretation Recall and Instance-level Recall suggests that models often recover only part of the expert interpretations, rather than covering it comprehensively.

\begin{figure*}[t]
  \centering
  \includegraphics[width=\textwidth]{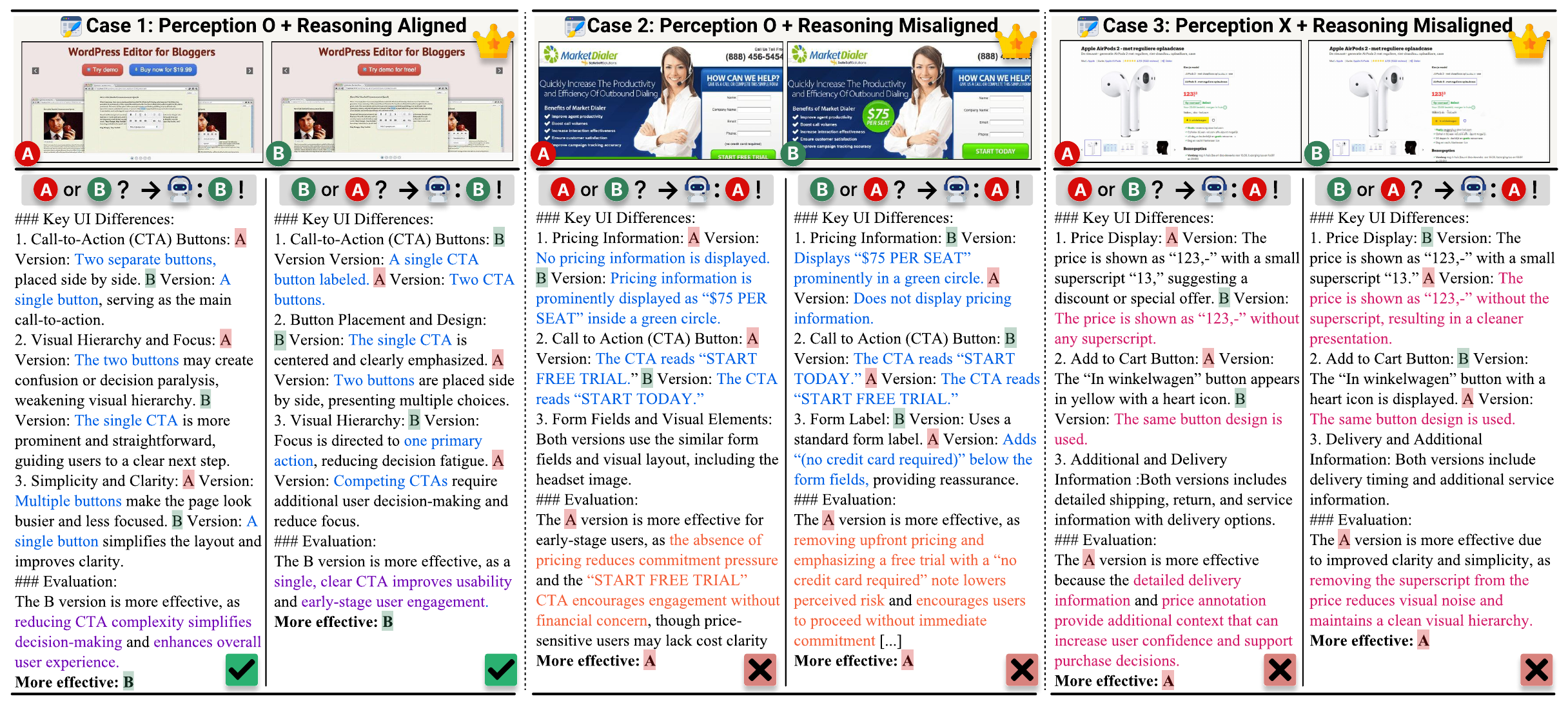}
    \caption{
    Example responses from InternVL-2.5-38B on the UI/UX design selection task, showing model outputs under different input image orders for each case. {\color[HTML]{0060E6}Blue} marks correct UI change perception, while {\color[HTML]{D31D69}red} denotes incorrect or hallucinated perception. {\color[HTML]{7209B7}Purple} indicates user behavior pattern-aligned reasoning, whereas misaligned reasoning in {\color[HTML]{F4613B}orange}.
    }
  \label{fig:quali_internvl}
\end{figure*}

\begin{figure*}[t]
  \centering
  \includegraphics[width=0.95\textwidth]{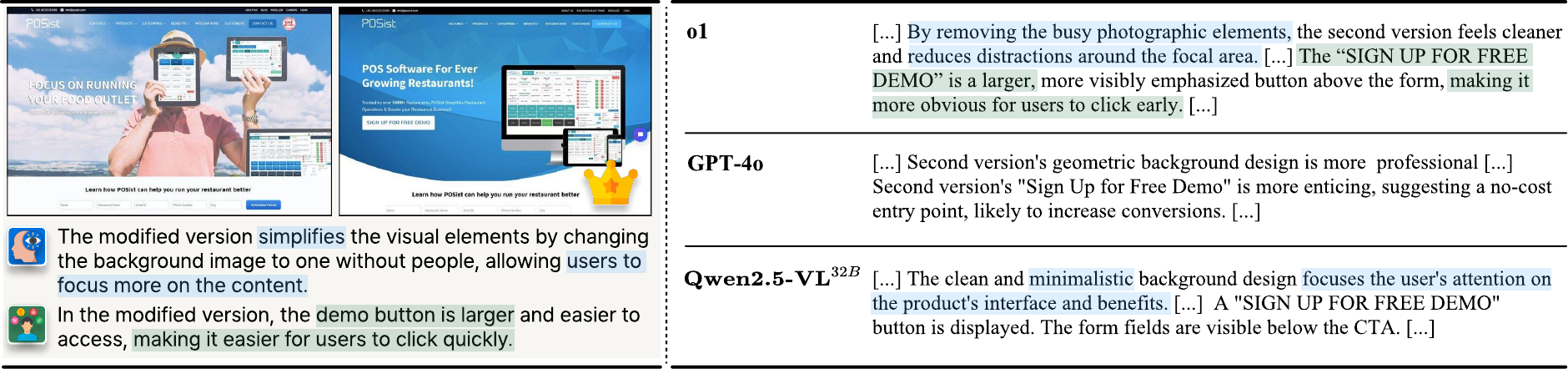}
  \caption{Example responses from o1, GPT-4o, and Qwen2.5-VL-32B on the UI/UX design interpretation task. {\color[HTML]{347F59}{Green}} and {\color[HTML]{5FB0FE}{blue}} highlight points capturing the expert interpretations fully.}
  \label{fig:quali}
\end{figure*}

\paragraph{Performance Across UX Aspects}

We further analyze Interpretation Recall scores by UX dimension to examine which types of UX interpretations are more readily recovered, as a true understanding of UI/UX requires capturing the full range of cognitive aspects—perception, memory, and action. As shown in Figure \ref{fig:ux_rationale}, o1 performs best on Perception, indicating relatively stronger ability to explain which visual cues are likely noticed by users, whereas GPT-5.1 achieves the highest score on Action, suggesting better performance on interaction-oriented explanations. More broadly, no single model dominates uniformly across all three dimensions. Performance remains below or around 70\% across most dimensions, underscoring the need for more holistic modeling of UI/UX understanding across multiple cognitive aspects.

\section{Discussion} \label{main:discussion}

\paragraph{Case Study}


For the UI/UX design selection task, we analyze the success and failure modes of the strong-performing model, InternVL-2.5-38B, as shown in Figure \ref{fig:quali_internvl}, focusing on cases where the model makes consistent decisions under both input image orders. In Case 1, the model correctly perceives the UI change as the presence of a CTA button and selects the more effective UI/UX, providing a reason that behaviorally aligns with the user behavior pattern implied by the A/B test outcome, hypothesizing that fewer CTAs better focus user attention. In Case 2, although the model correctly identifies differences such as pricing badges and CTA text, it fails to select the winning variant identified by the A/B test, instead offering misaligned reasoning that higher visual saliency negatively impacts user behavior by inducing commitment pressure. In Case 3, the model fails at perception level, hallucinating non-existent changes instead of recognizing the actual button color difference, which ultimately leads to an incorrect final decision.

Figure \ref{fig:quali} shows a case from the UI/UX design interpretation task. The UI on the right was verified by A/B test results as the more effective design, with experts attributing its effectiveness to a simplified background and an enlarged demo button, along with their associated effects, factors that models must explicitly capture for an accurate interpretation. o1 succeeds by directly highlighting these key points. GPT-4o correctly identifies the relevant UI elements, but provides an incorrect or vague explanation for their impact. Qwen2.5-VL-32B correctly notes the minimalist design, but stops at identifying the button, missing the key detail of its enlargement and resulting in an incomplete explanation. As shown here, models often identify and emphasize salient surface-level differences, but fall short in capturing deeper behavioral implications and UX reasoning, limiting interpretive depth relative to experts.

\section{Conclusion}

We introduce a new task that evaluates multimodal understanding of how design drives user actions in the critical field of UI/UX. To support this, we present \benchmark, a behavior-focused benchmark uniquely built from diverse, real-world A/B‑tested UI variants and expert-curated key interpretations, providing a reliable basis for assessment. Our two tasks, selection and interpretation, go beyond surface-level recognition, requiring models to reason about the behavioral implications of visual design, a cognitively challenging problem. Extensive experiments across both proprietary and open-source MLLMs reveal they exhibit only shallow levels of behavior-aware visual reasoning. Achieving stronger performance on this significant task, particularly across the analyzed UI elements or UX dimensions, may require further training on large-scale, behaviorally grounded UI/UX data that reflects real-world user dynamics. We envision \benchmark and our work as a step toward multimodal systems capable of not only understanding but also generating visual interfaces grounded in user behavior contexts, aligning with human cognition to optimize user experience. More broadly, we hope to catalyze future research toward deeper UI/UX and human behavior literacy in AI.

\section{Limitations}

Given that user experience patterns vary across cultural contexts and societal norms, cultural biases inevitably exist in our benchmark \citep{karreman2016cross}.
Also, interactive UIs were not extensively explored in our work. Although the number of 300 UI image pairs may seem limited, publicly available real-world A/B test results are extremely rare, as companies rarely disclose them. Moreover, independently conducting production-level UI experiments at a comparable level and scale was practically infeasible. However, as the first benchmark in this area, we focused on reliability, diversity, and annotation quality, covering a wide range of real-world UI changes and providing expert-curated key interpretations.


\section{Ethical Considerations}

To ensure expert-level annotation quality, we recruited three annotators with professional UI/UX backgrounds who are currently employed at global IT companies specializing in UX and AI. We clearly informed them of the academic nature of the task, and they voluntarily participated in the study. We provided each annotator with monetary compensation equivalent to approximately \$225 USD, reflecting fair market rates for domain-specific labor, and issued payments within 48 hours of annotation completion. We instructed annotators to focus solely on task-relevant elements and ensured that they were not exposed to or asked to label any sensitive, offensive, or triggering content.


Our benchmark does not include any private, user-identifiable, or confidential information. All materials were curated exclusively from publicly accessible promotional case studies (e.g., \textit{Success Stories} from VWO\footnote{\url{https://vwo.com/success-stories}}) released by the originating platforms, while strictly excluding any restricted or paywalled content. We acknowledge that while some source platforms include copyright notices intended to protect their proprietary assets, they do not provide explicit open redistribution licenses, such as Creative Commons, for their publicly accessible case materials. To responsibly address the absence of explicit licenses and to ensure compliant academic use, \benchmark is released strictly for non-commercial academic research under a non-commercial license (e.g., CC BY-NC-SA 4.0). Rather than redistributing raw image assets directly, we primarily release source URLs alongside our expert annotations and automated image-processing scripts. We will further maintain transparent communication channels within the repository to accommodate potential inquiries from original content owners and, if necessary, adjust specific entries in accordance with responsible academic data governance.


\section*{Acknowledgments}
This work was supported by the Institute of Information \& Communications Technology Planning \& Evaluation (IITP) grant funded by the Korea Government (MSIT) (No. RS-2024-00338140, Development of learning and utilization technology to reflect sustainability of generative language models and up-to-dateness over time), and by the Culture, Sports and Tourism R\&D Program through the Korea Creative Content Agency grant funded by the Ministry of Culture, Sports and Tourism in 2026 (RS-2024-00361757, Development of Multimodal UX Evaluation Platform Technology for XR Spatial Responsive Content Optimization)


\bibliography{custom}

\clearpage

\appendix

\input{supple}

\end{document}

%% file: supple.tex
\section{\benchmark Details} 

\subsection{Curation Details} \label{appx:annotation}

\paragraph{Image Preprocess}

\begin{figure*}[h]
  \centering
  \includegraphics[width=0.7\textwidth]{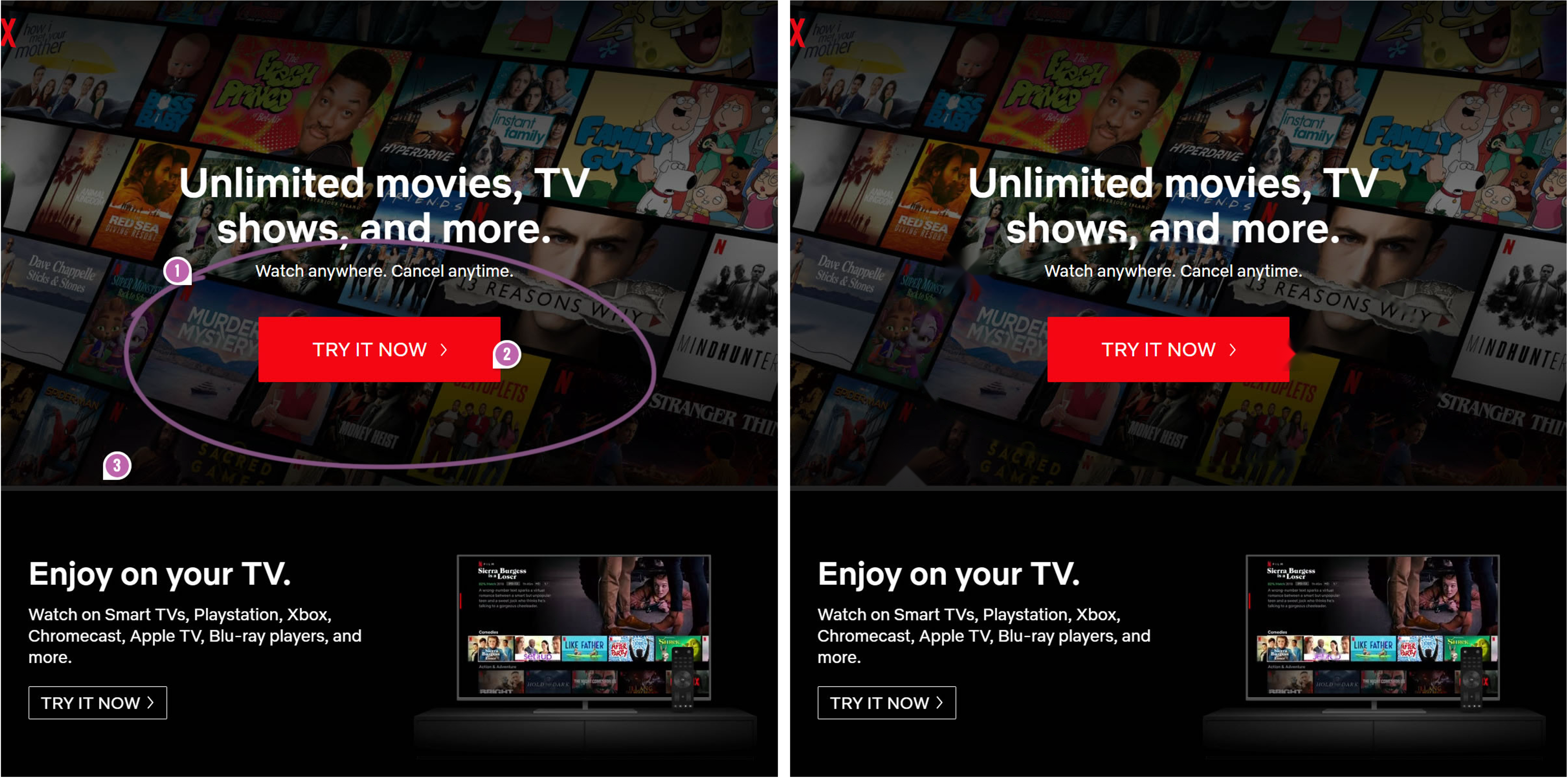}
  \caption{Example of visual indicator removal in \benchmark. Left: Original image with markers added by data source website. Right: After inpainting, with markers removed while preserving UI elements.}
  \label{fig:inpainting_example}
\end{figure*}

During benchmark construction process, we observed that some UI images from our data sources included manual indicators, such as circled regions, numbered markers, and arrows, highlighting specific UI elements related to reasoning points from A/B test results (see Figure \ref{fig:inpainting_example}). While helpful for human interpretation, these indicators introduced artificial attention cues that could potentially bias model evaluation.

To mitigate this, we cleaned the dataset using inpainting techniques. We first identified common indicator patterns via edge detection and color-based segmentation, including circular highlights drawn in semi-transparent colors, numbered markers, and arrows. Our automated pipeline primarily utilized OpenCV \citep{opencv_library} for marker detection and inpainting. For a small number of complex cases where automated methods were insufficient, we applied a hybrid refinement approach using AI-based editing tools (e.g., Gemini\footnote{\url{https://gemini.google.com}}). These tools were used solely to remove attached markers or visual indicators, without modifying the underlying UI layout or design elements. This process ensured a cleaner, more naturalistic dataset for fair visual reasoning evaluation.

\paragraph{Interpretation Annotation}

\begin{figure*}[t]
  \centering
  \includegraphics[width=0.8\textwidth]{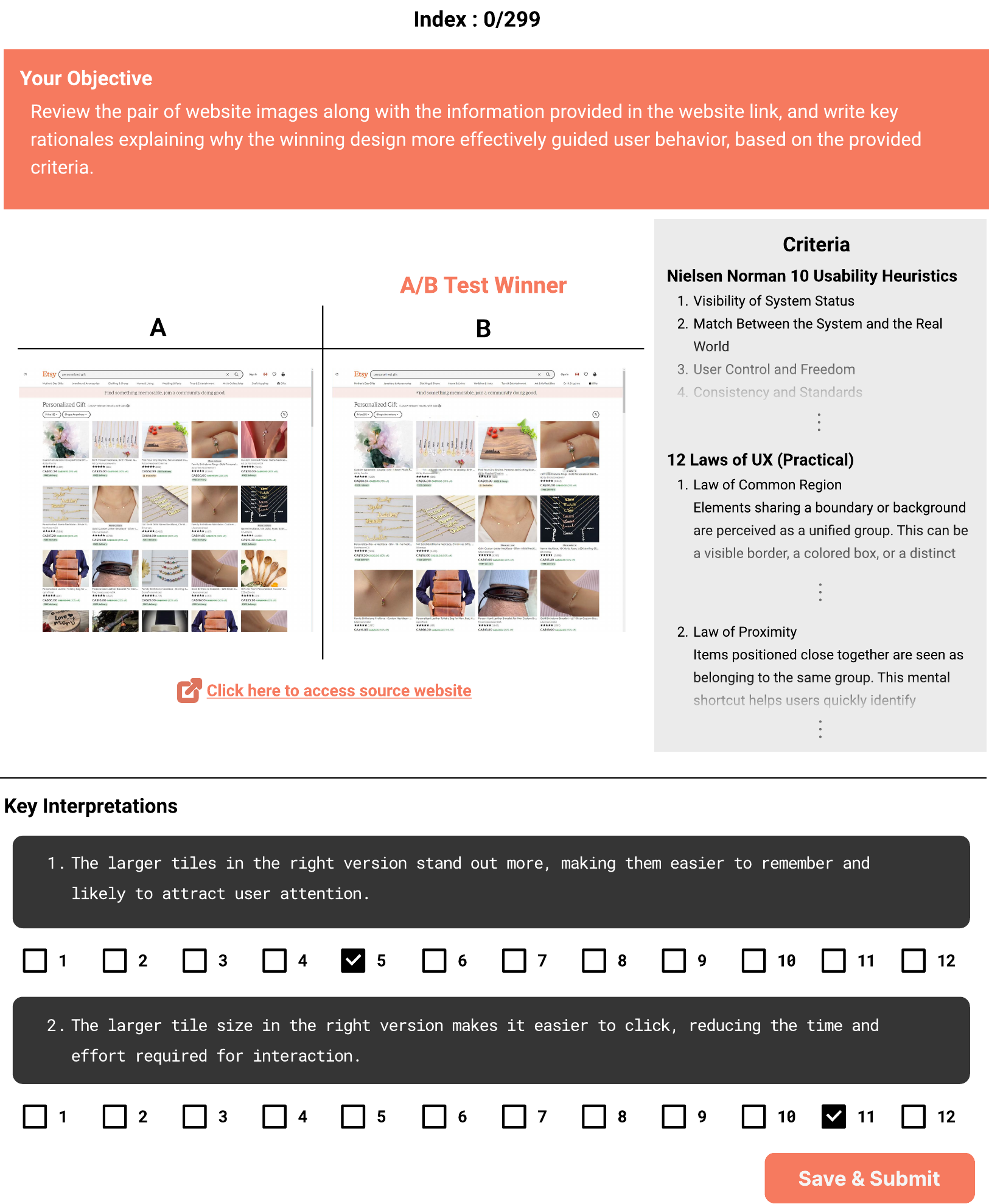}
  \caption{A screenshot of the interpretation annotation interface for UI/UX experts based on original data sources and guidelines. After writing the key interpretations, they also selected the most relevant UX law.}
  \label{fig:expert_anno}
\end{figure*}

As shown in Figure \ref{fig:expert_anno}, three UI/UX experts were presented with a pair of UI images, one designated as the winner based on A/B test results, along with the corresponding source website link. Experts were instructed to provide key interpretations explaining why the winning variant supported user actions more effectively than the alternative. We engaged domain experts for this task, as user behavior is often subconscious and shaped by implicit context, making evaluation by non-experts inherently difficult.

Annotators were guided by two resources: textual descriptions from the original sources, which often highlight key design intentions, and the Nielsen Norman 10 Usability Heuristics \citep{nielsen1994enhancing}, a robust theoretical framework widely used in usability research, which served as broad conceptual criteria. The resulting interpretations were then mapped to 12 curated UX laws that remain widely applied in practice. These laws provide concrete, actionable guidance across diverse aspects of user experience, which can be categorized into three cognitive dimensions: Perception, Memory, and Action. This classification is inspired by the framework proposed by \citet{norman2007emotional}, which introduced the concept of \textit{Emotional Design}. The full list of UX laws and their classifications is presented in Table \ref{tab:apdx_benchmark_detail}.




\subsection{Statistics Details} \label{appx:stat}

\begin{figure*}[t]
\centering
\includegraphics[width=0.9\textwidth]{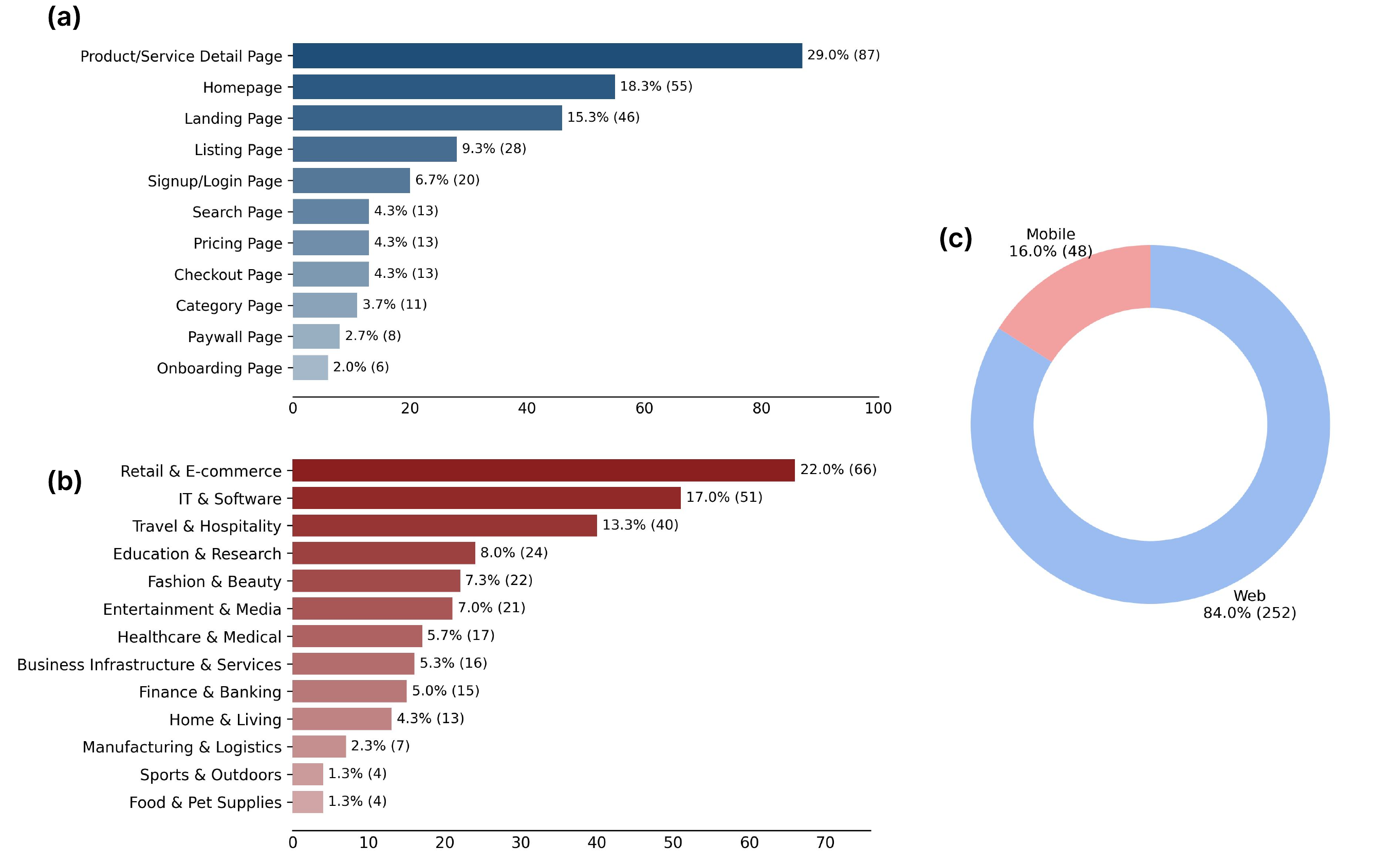}
\caption{Detailed statistics of \benchmark on (a) page type, (b) industry domain, (c) device type}
\label{fig:more_stat}
\end{figure*}

Figure~\ref{fig:more_stat} presents the distribution of \benchmark across page types, industry domains, device types. For clarification on the page type classification, the \textit{homepage} refers to the main entry point of a website, typically offering broad navigation and general content. In contrast, a \textit{landing page} is usually a standalone page optimized for a specific user action, such as signing up or making a purchase.

\begin{figure*}[t]
\centering
\includegraphics[width=0.6\textwidth]{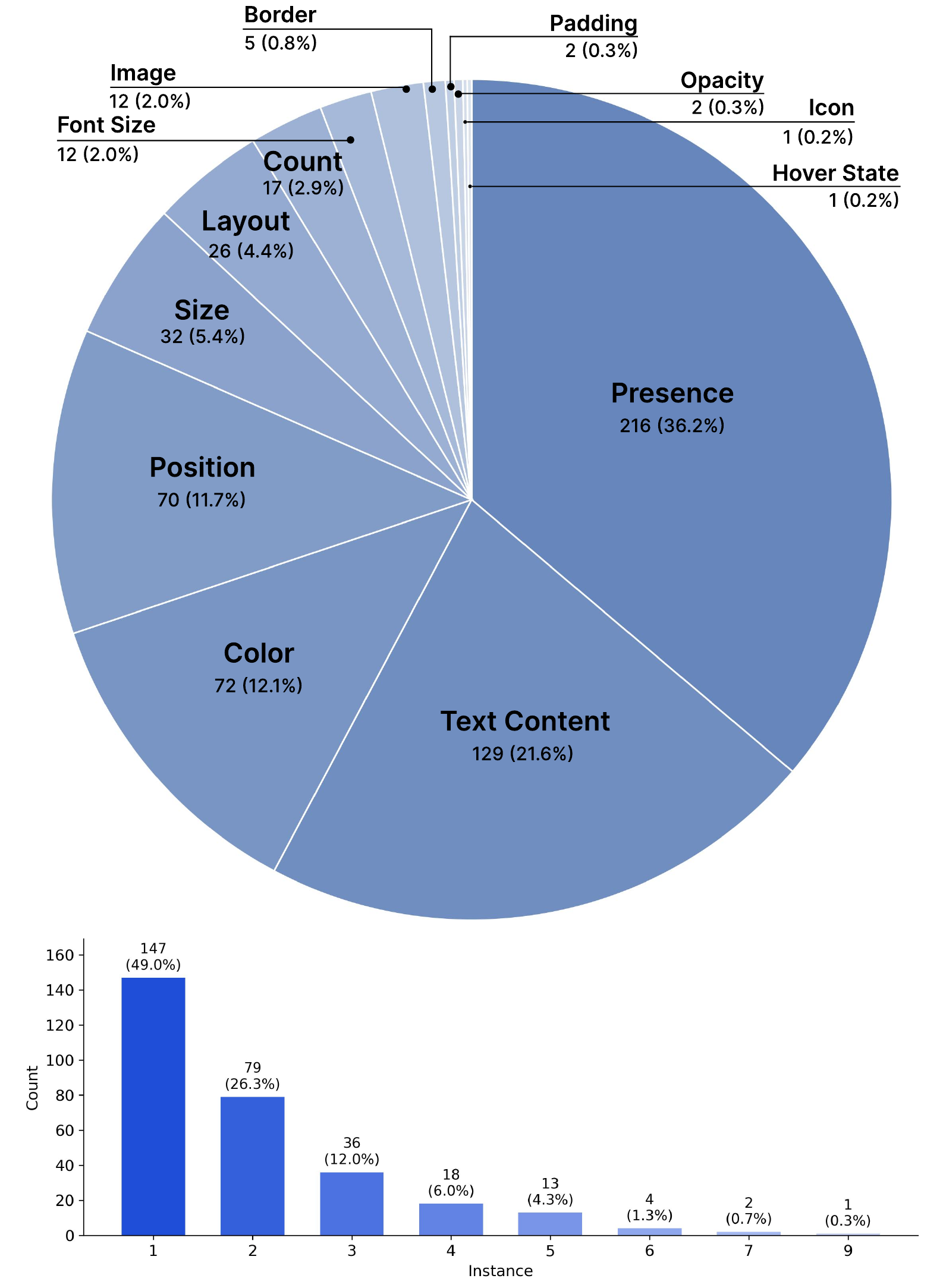}
\caption{Detailed statistics of \benchmark on UI change attribute type}
\label{fig:more_stat_modi}
\end{figure*}

Full distribution of UI change attribute types is shown in Figure \ref{fig:more_stat_modi}. For clarification, element-level counts reflect the number of distinct UI components affected, while attribute-level counts capture the total number of modified attributes across all components. For example, an instance annotated as <Input Field: Padding, Color / Button: Color> is counted as 2 element changes and 3 attribute changes.

\section{Experiment Details} \label{appx:exp}

\subsection{Setup} \label{appx:setup}
\subsubsection{Model Configuration}

For proprietary MLLMs, we accessed the models via their official API endpoints as follows:

\begin{itemize}
    \item o1 \cite{openai2024openaio1card}:\\ \texttt{o1-2024-12-17}
    \item GPT-5.1 \cite{openai2025gpt51}:\\ \texttt{gpt-5.1-2025-11-13} \\ (\texttt{reasoning effort}: none)
    \item GPT-4o \cite{openai2024gpt4ocard}: \\ \texttt{gpt-4o-2024-08-06}
    \item Claude 4.5 Sonnet \cite{anthropic2025claudesonnet45}:\\ \texttt{claude-sonnet-4-5-20250929} \\ (\texttt{thinking type}: disabled)
    \item Claude 3.5 Sonnet \cite{anthropic2024claude3}:\\ \texttt{claude-3-5-sonnet-20240620}
\end{itemize}

For open-source models, we used publicly available weights from the Huggingface Hub\footnote{https://huggingface.co/models}. Specific model versions are as follows:

\begin{itemize}
    \item Qwen3-VL-30B-A3B \cite{bai2025qwen3}:\\ \texttt{Qwen/Qwen3-VL-30B-A3B-Instruct}
    \item Qwen3-VL-8B \cite{bai2025qwen3}:\\ \texttt{Qwen/Qwen3-VL-8B-Instruct}
    \item Qwen2.5-VL-32B \cite{bai2025qwen25vltechnicalreport}:\\ \texttt{Qwen/Qwen2.5-VL-32B-Instruct}
    \item Qwen2.5-VL-7B \cite{bai2025qwen25vltechnicalreport}:\\ \texttt{Qwen/Qwen2.5-VL-7B-Instruct}
    \item InternVL-2.5-38B \cite{chen2025expandingperformanceboundariesopensource}: \\\texttt{OpenGVLab/InternVL2\string_5-38B}
    \item InternVL-2.5-8B \cite{chen2025expandingperformanceboundariesopensource}:\\ \texttt{OpenGVLab/InternVL2\string_5-8B}
    \item LLaVA-NeXT-7B \cite{liu2024improved}: \\\texttt{llava-hf/llava-v1.6-mistral-7b-hf}
    \item LLaVA-OneVision-7B \cite{li2024llavaonevisioneasyvisualtask}:\\ \texttt{llava-hf/llava-onevision-qwen2-7b-ov-\allowbreak{}hf}
\end{itemize}

The temperature for all models was set to 0.2.

\subsubsection{Resources}

All experiments using open-source models were conducted on a Linux machine (Ubuntu 20.04.6 LTS) equipped with 4 NVIDIA RTX A6000 GPUs, each with 48GB VRAM. We used the following software libraries and frameworks: \texttt{transformers} version 4.53.2 and \texttt{vllm} version 0.9.2.


\subsection{UI/UX Design Selection}  \label{appx:selection}

\subsubsection{Prompts}
Figures \ref{prompt:zero_shot} through \ref{prompt:mad_moderator_extractive} show the prompts used, including those designed for diverse reasoning strategies.

\subsubsection{Token Counts}

Token counts across reasoning strategies on proprietary models are presented in Table \ref{tab:token}.

\subsubsection{Standard Deviation}
Table \ref{tab:std_analysis} presents the standard deviation across three experiment runs for each model and reasoning method. For metric abbreviation, \textit{FS} and \textit{SS} denote the standard deviation when the first or second UI image input was the more effective one, while \textit{AS} and \textit{CS} refer to the standard deviation of total correct and both correct cases, respectively.

\subsubsection{Additional Case Study}
\begin{figure*}[h]
    \centering
    \includegraphics[width=0.95\textwidth]{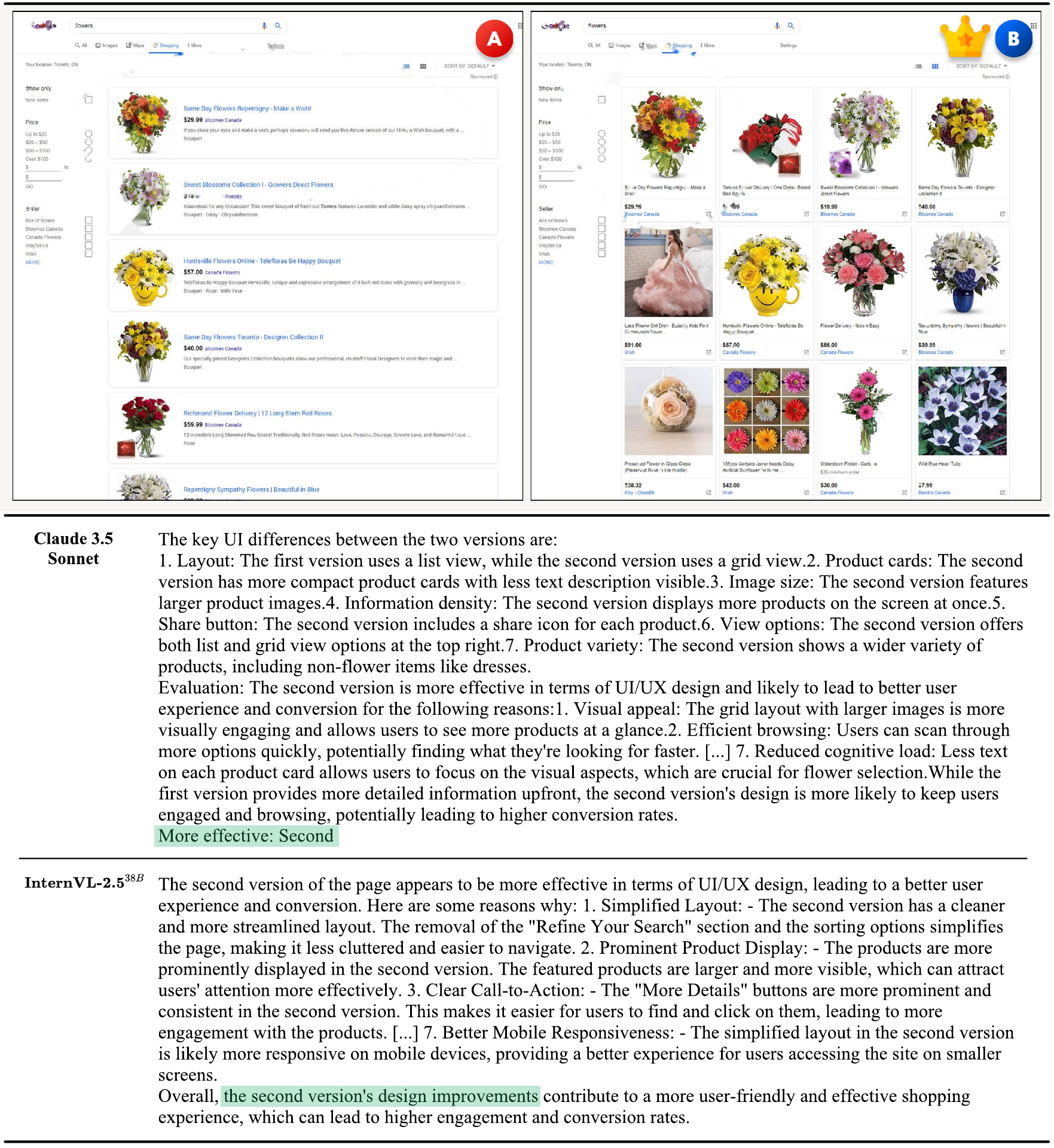}
    \caption{A qualitative example from the UI/UX design selection task on \benchmark: Both Claude 3.5 Sonnet and InternVL-2.5-38B selected the A/B test winner.}
    \label{fig:appendix_quali_task1_1}
\end{figure*}

\begin{figure*}[h]
    \centering
    \includegraphics[width=0.95\textwidth]{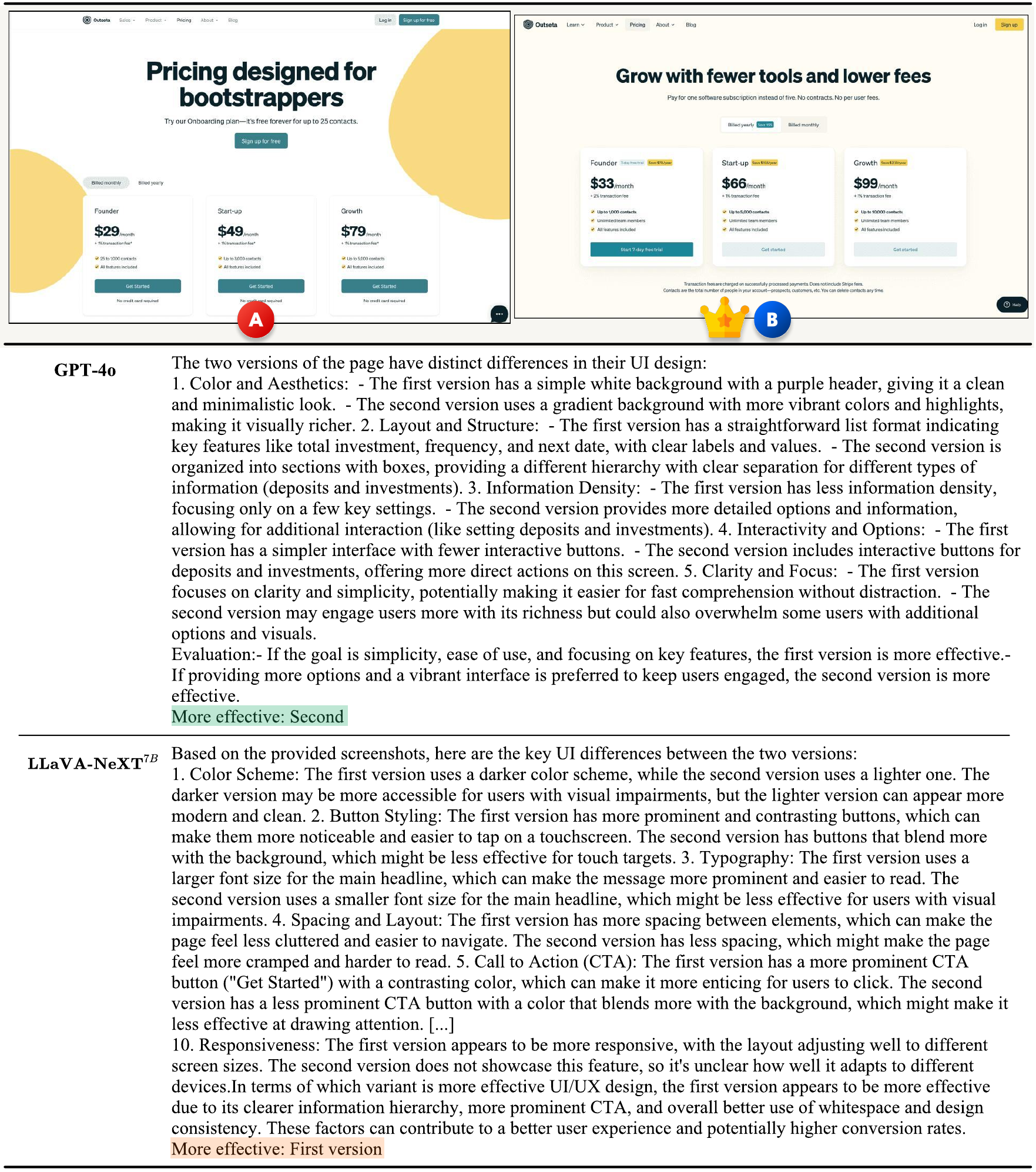}
    \caption{A qualitative example from the UI/UX design selection task on \benchmark: GPT-4o selected the A/B test winner, whereas LLaVA-NeXT-7B did not.}
    \label{fig:appendix_quali_task1_2}
\end{figure*}

\begin{figure*}[h]
    \centering
    \includegraphics[width=0.92\textwidth]{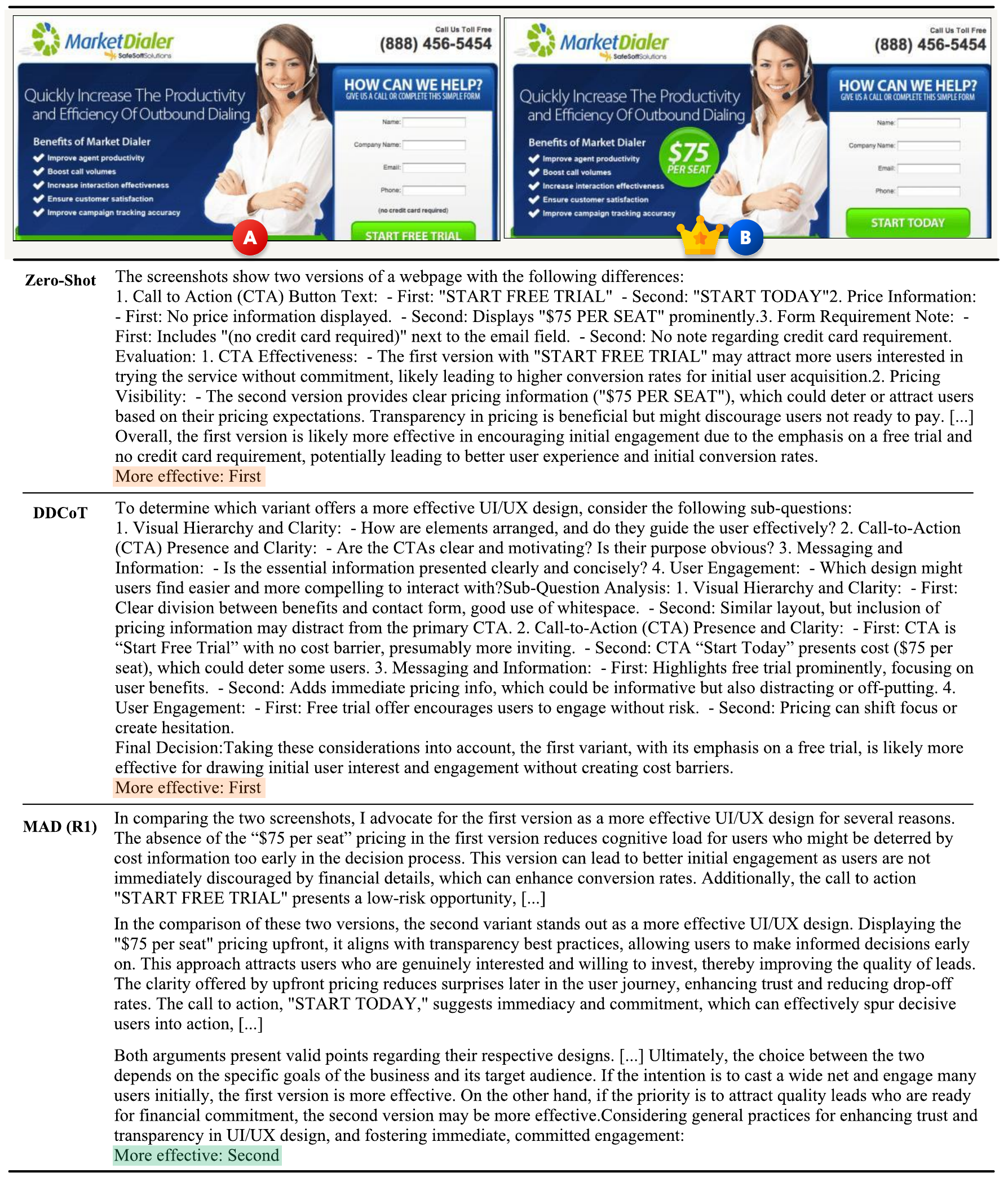}
    \caption{A qualitative example from the UI/UX design selection task on \benchmark, showcasing the effect of diverse reasoning strategies executed on GPT-4o.}
    \label{fig:appendix_quali_reasoning}
\end{figure*}

Figure \ref{fig:appendix_quali_task1_1} and \ref{fig:appendix_quali_task1_2} present the additional case studies across different models. Figure \ref{fig:appendix_quali_reasoning} shows an example comparing various reasoning strategies executed with GPT-4o.

\subsection{UI/UX Design Interpretation} \label{appx:interpret}

\subsubsection{Prompts}

Figure \ref{prompt:task2_zero_shot} presents the prompt used for models' interpretation generation, and Figure \ref{prompt:task2_gpt_eval} presents the prompt used for its GPT-4o-based evaluation.

\subsubsection{Results on Text Similarity Metrics}

Alongside Interpretation Recall and Instance-level Recall from GPT-4o-based evaluation, we report BLEU, ROUGE, METEOR, and BERTScore \cite{papineni-etal-2002-bleu, lin-2004-rouge, banerjee-lavie-2005-meteor, zhang2020bertscoreevaluatingtextgeneration} in Table \ref{tab:text_similarity} as complementary lexical-overlap metrics.

The consistently low scores observed in BLEU, ROUGE, and METEOR primarily stem from verbosity and vocabulary differences between concise expert interpretations and more elaborative model-generated explanations, as also illustrated in Figure \ref{fig:quali}. As such, these metrics largely capture stylistic divergence rather than true semantic misalignment, In contrast, our recall-based metrics are explicitly designed to assess semantic inclusion, thereby providing a more faithful characterization of the alignment between expert and model interpretations, which motivates their use as the main evaluation criterion.


\subsubsection{Additional Case Study}

\begin{figure*}[h]
    \centering
    \includegraphics[width=0.95\textwidth]{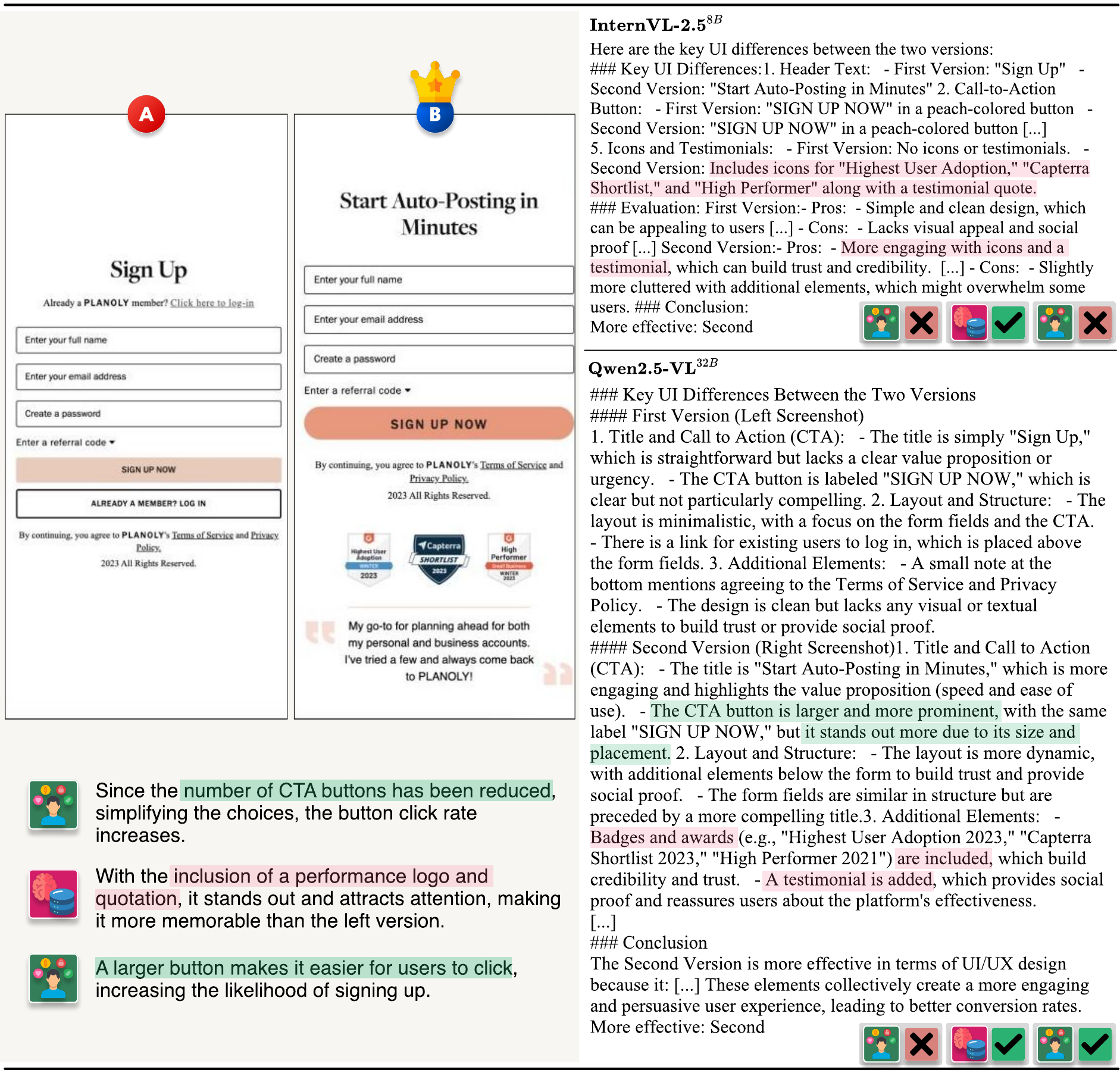}
    \caption{A qualitative example from the UI/UX design interpretation task on \benchmark: InternVL-2.5-8B captured only the memory part, while Qwen2.5-VL-32B identified two key interpretations. The highlights in the models' responses indicate the parts that align with the expert-curated key interpretations.}
    \label{fig:appendix_quali_task2}
\end{figure*}

Figure \ref{fig:appendix_quali_task2} presents an additional case study across different models.



\section{Icon Attribution}

Icons used in this paper are from Flaticon (https://www.flaticon.com), and they are attributed to the respective authors as required by Flaticon's license.

\clearpage

\begin{table*}[!t]
\centering
\resizebox{\linewidth}{!}{
\newcolumntype{C}[1]{>{\centering\arraybackslash}m{#1}}   
\newcolumntype{L}[1]{>{\raggedright\arraybackslash}p{#1}} 

\begin{tabular}{c|C{4cm}|L{9cm}}
\toprule
UX Dimension & UX Law & Description \\
\midrule
\multirow{13}{*}{Perception}
& \raisebox{-.5\height}{\parbox[c][3\baselineskip][c]{3.8cm}{\centering Law of Common Region\\\cite{koffka1922perception}}}
& Elements sharing a boundary or background are perceived as a unified group. This can be a visible border, a colored box, or a distinct region in the layout. \\
\cmidrule{2-3}
& \raisebox{-.5\height}{\parbox[c][3\baselineskip][c]{3.8cm}{\centering Law of Proximity\\\citep{koffka1922perception}}}
& Items positioned close together are seen as belonging to the same group. This mental shortcut helps users quickly identify relationships between adjacent elements. \\
\cmidrule{2-3}
& \raisebox{-.5\height}{\parbox[c][3\baselineskip][c]{3.8cm}{\centering Law of Prägnanz\\\citep{koffka1922perception}}}
& Humans naturally perceive and interpret complex images in their simplest forms. It reduces cognitive effort by helping us group and understand elements more quickly. \\
\cmidrule{2-3}
& \raisebox{-.5\height}{\parbox[c][3\baselineskip][c]{3.8cm}{\centering Law of Similarity\\\citep{koffka1922perception}}}
& People group together elements that look alike, assuming they serve a similar purpose. Consistency in color, shape, or size leads the user to see these items as related. \\
\midrule
\multirow{13}{*}{Memory}
& \raisebox{-.5\height}{\parbox[c][3\baselineskip][c]{3.8cm}{\centering Von Restorff Effect\\\citep{von1933wirkung}}}
& People remember a distinct element more easily than others that blend in. This is also known as the Isolation Effect, emphasizing our attention on anything unusual. \\
\cmidrule{2-3}
& \raisebox{-.5\height}{\parbox[c][3\baselineskip][c]{3.8cm}{\centering Serial Position Effect\\\citep{ebbinghaus1913memory}}}
& Users tend to remember the first and last items in a series more than those in the middle. This influences how they recall sequential information or actions. \\
\cmidrule{2-3}
& \raisebox{-.5\height}{\parbox[c][3\baselineskip][c]{3.8cm}{\centering Miller's Law\\\citep{miller1956magical}}}
& On average, people can hold only about seven items in their working memory at once. Going beyond this limit causes cognitive overload. \\
\cmidrule{2-3}
& \raisebox{-.5\height}{\parbox[c][3\baselineskip][c]{3.8cm}{\centering Jakob's Law\\\citep{nielsen2000end}}}
& Users expect new interfaces to operate similarly to those they've used before. Leveraging established patterns lowers cognitive load and accelerates adoption. \\
\midrule
\multirow{13}{*}{Action}
& \raisebox{-.5\height}{\parbox[c][3\baselineskip][c]{3.8cm}{\centering Hick's Law\\\citep{hick1952rate}}}
& Decision time rises with the number and complexity of choices. Too many options can overwhelm users and slow their actions. \\
\cmidrule{2-3}
& \raisebox{-.5\height}{\parbox[c][3\baselineskip][c]{3.8cm}{\centering Pareto Principle\\\citep{pareto1919manuale}}}
& Roughly 80\% of outcomes stem from 20\% of causes. Focusing on the most impactful features or tasks delivers the greatest improvement. \\
\cmidrule{2-3}
& \raisebox{-.5\height}{\parbox[c][3\baselineskip][c]{3.8cm}{\centering Fitts's Law\\\citep{fitts1964information}}}
& The time required to reach a target depends on its size and distance. Larger, closer elements are easier to access quickly and accurately. \\
\cmidrule{2-3}
& \raisebox{-.5\height}{\parbox[c][3\baselineskip][c]{3.8cm}{\centering Goal-Gradient Effect\\\citep{hull1932goal}}}
& Motivation increases as people get closer to finishing a task. Providing clear progress markers encourages users to continue and complete it faster. \\
\bottomrule
\end{tabular}}
\caption{Overview of UX laws used for interpretation annotation, and their categorization by cognitive dimension (Perception, Memory, Action).}
\label{tab:apdx_benchmark_detail}
\end{table*}

\begin{table*}[h]
\centering
\begin{tabular}{c|c|cc}
\toprule
Models & Methods & Input Tokens & Output Tokens \\
\midrule

\multirow{1}{*}{o1} & Zero-Shot & 1292.5 ($\pm$ 559.77) & 1347.7 ($\pm$ 489.76) \\

\midrule

\multirow{6}{*}{GPT-4o} & Zero-Shot & 1453.4 ($\pm$ 634.41) & 248.8 ($\pm$ 57.98) \\
                        & CoCoT & 1458.4 ($\pm$ 634.41) & 329.7 ($\pm$ 63.29)  \\
                        & Self-Refine & 5240.4 ($\pm$ 1935.80) & 820.9 ($\pm$ 142.43)  \\
                        & DDCoT & 1521.4 ($\pm$ 634.41) & 415.3 ($\pm$ 73.42) \\
                        & MAD (R1) & 4942.4 ($\pm$ 1903.15) & 483.1 ($\pm$ 58.5) \\
                        & MAD (R3) & 11501.5 ($\pm$ 4435.48) & 1158.1 ($\pm$ 82.60) \\
\midrule
\multirow{6}{*}{\shortstack{Claude 3.5 \\ Sonnet}} & Zero-Shot & 1686.6 ($\pm$ 1062.43) & 401.4 ($\pm$ 72.94) \\
                        & CoCoT & 1692.5 ($\pm$ 1062.58) & 487.0 ($\pm$ 82.28) \\
                        & Self-Refine & 6230.7 ($\pm$ 3245.69) & 1201.4 ($\pm$ 207.32) \\
                        & DDCoT & 1754.5 ($\pm$ 1064.29) & 582.2 ($\pm$ 97.13) \\
                        & MAD (R1) & 5895.5 ($\pm$ 3224.31) & 834.1 ($\pm$ 125.13) \\
                        & MAD (R3) & 13853.4 ($\pm$ 7512.63) & 1936.5 ($\pm$ 276.14) \\
\bottomrule
\end{tabular}
\caption{Overview of Token Counts ($\mu\pm\sigma$) in the UI/UX design selection task on \benchmark.}
\label{tab:token}
\end{table*}

\begin{table*}[ht]
\centering
\begin{tabular}{l|c|cc|cc}
\toprule
Model & Method & FS & SS & AS & CS \\
\midrule
o1      & Zero-Shot                  & 1.528  & 2.082  & 3.606  & 0.577 \\
\midrule
\multirow{6}{*}{GPT-4o} 
        & Zero-Shot                  & 1.528  & 1.000  & 2.082  & 1.528 \\
        & CoCoT                         & 4.000  & 2.646  & 1.732  & 1.732 \\
        & Self-Refine                & 2.517  & 1.000  & 3.215  & 0.577 \\
        & DDCoT                       & 2.082  & 2.309  & 0.577  & 2.517 \\
        & MAD (R1) & 3.215  & 3.055  & 6.245  & 2.646 \\
        & MAD (R3) & 0.577  & 2.082  & 1.732  & 3.215 \\
\midrule
\multirow{6}{*}{Claude 3.5 Sonnet} 
        & Zero-Shot                  & 1.528  & 2.000  & 3.512  & 2.887 \\
        & CoCoT                         & 3.055  & 3.512  & 0.577  & 0.577 \\
        & Self-Refine                & 2.646  & 3.512  & 4.619  & 1.528 \\
        & DDCoT                       & 2.082  & 1.732  & 0.577  & 2.517 \\
        & MAD (R1) & 1.000  & 1.528  & 2.309  & 4.359 \\
        & MAD (R3) & 1.000  & 2.887  & 3.786  & 3.786 \\
\midrule
\shortstack{Qwen2.5-VL-32B}
& Zero-Shot& 1.732 & 2.309 & 2.082 & 2.000 \\
\shortstack{Qwen2.5-VL-7B}
& Zero-Shot& 2.646 & 0.577 & 2.517 & 3.215 \\
InternVL-2.5-38B
& Zero-Shot& 2.646 & 1.000 & 3.000 & 3.055 \\
InternVL-2.5-8B
& Zero-Shot& 1.000 & 2.646 & 2.646 & 2.000 \\
LLaVA-NeXT-7B
& Zero-Shot& 3.512 & 1.155 & 4.583 & 6.028 \\
LLaVA-OneVision-7B
& Zero-Shot& 4.041 & 2.309 & 4.359 & 3.512 \\
\bottomrule
\end{tabular}
\caption{Results on standard deviation across three experiment runs in the UI/UX design selection task for each model and reasoning strategy on \benchmark.}
\label{tab:std_analysis}
\end{table*}

\begin{table*}[t]
\centering
\begin{tabular}{l|ccccc}
\toprule
Model & BLEU-2 & BLEU-4 & ROUGE-L & METEOR & BERTScore \\
\midrule
o1 & 0.0181 & 0.0041 & 0.0972 & 0.1630 & 0.8414 \\
GPT-4o & 0.0162 & 0.0040 & 0.0875 & 0.1557 & 0.8233 \\
Claude 3.5 Sonnet & 0.0141 & 0.0037 & 0.0654 & 0.1418 & 0.8076 \\
\midrule
Qwen2.5-VL-32B & 0.0098 & 0.0025 & 0.0442 & 0.1016 & 0.8085 \\
Qwen2.5-VL-7B & 0.0180 & 0.0048 & 0.0789 & 0.1481 & 0.8331 \\
InternVL-2.5-38B & 0.0128&0.0034&0.0574&0.1225&0.8080 \\
InternVL-2.5-8B & 0.0111 & 0.0032 & 0.0510 & 0.1124 & 0.8082 \\
LLaVA-NeXT-7B &0.0077&0.0018&0.0386&0.0908&0.8127 \\
LLaVA-OneVision-7B &0.0157&0.0043&0.0718&0.1416&0.8255 \\

\bottomrule
\end{tabular}
\caption{Results of text similarity metrics per expert interpretation for the UI/UX design interpretation task on \benchmark.}
\label{tab:text_similarity}
\end{table*}





\clearpage

\begin{figure}[b]
\centering
\begin{tcolorbox}[
  colback=gray!5!white,
  colframe=gray!75!black,
  arc=2mm, %
    fontupper=\small
]
You are an expert in designing UI/UX for web/apps. \\

The two screenshots show two different versions of the same page.\\
Identify the key UI differences between the two versions, and then evaluate which variant is more effective UI/UX design that leads to better user experience and conversion.\\

You should end your answer with following the format (No bold, etc): \\
More effective: \textless{}First/Second\textgreater{}

\end{tcolorbox}
\caption{Prompt used for the UI/UX design selection task with zero-shot strategy.}
\label{prompt:zero_shot}
\end{figure}

\begin{figure}[b]
\centering
\begin{tcolorbox}[
  colback=gray!5!white,
  colframe=gray!75!black,
  arc=2mm,
  fontupper=\small
]
You are an expert in designing UI/UX for web/apps. \\

The two screenshots show two different versions of the same page.  \\
Identify the key UI differences between the two versions, and then evaluate which variant is more effective UI/UX design that leads to better user experience and conversion. \\

Think step by step. \\

You should end your answer with the following format (No bold, etc):  \\
More effective: \textless{}First/Second\textgreater{}
\end{tcolorbox}
\caption{Prompt used for the UI/UX design selection task with CoCoT strategy.}
\label{prompt:cot}
\end{figure}

\begin{figure}[b]
\centering
\begin{tcolorbox}[
  colback=gray!5!white,
  colframe=gray!75!black,
  arc=2mm,
  fontupper=\small
]
You are an expert in designing UI/UX for web/apps. \\

The two screenshots show two different versions of the same page.\\
We want to identify the key UI differences between the two versions, and then evaluate which variant is more effective UI/UX design that leads to better user experience and conversion.\\

Review your previous answer and find problems with your answer.\\
Previous answer:\\
\textbf{\{previous\_answer\}} \\

\end{tcolorbox}
\caption{Prompt used for the UI/UX design selection task with Self-Refine strategy on the reviewing step. Prompt inputs are in \textbf{bold}.}
\label{prompt:self_refine_2_review}
\end{figure}

\begin{figure}[b]
\centering
\begin{tcolorbox}[
  colback=gray!5!white,
  colframe=gray!75!black,
  arc=2mm,
  fontupper=\small
]
You are an expert in designing UI/UX for web/apps.\\

The two screenshots show two different versions of the same page.\\
We want to identify the key UI differences between the two versions, and then evaluate which variant is more effective UI/UX design that leads to better user experience and conversion.\\

Your previous answer:\\
\textbf{\{previous\_answer\}} \\

Feedback on your previous answer:\\
\textbf{\{feedback\}} \\

Based on the feedback for your previous answer, improve your answer.\\

You should end your answer with following the format (No bold, etc):\\
More effective: \textless{}First/Second\textgreater{}\\
\end{tcolorbox}
\caption{Prompt used for the UI/UX design selection task with Self-Refine strategy on the improving step. Prompt inputs are in \textbf{bold}.}
\label{prompt:self_refine_3_improve}
\end{figure}

\begin{figure}[b]
\centering
\begin{tcolorbox}[
  colback=gray!5!white,
  colframe=gray!75!black,
  arc=2mm,
  fontupper=\small
]
You are an expert in designing UI/UX for web/apps. \\

The two screenshots show two different versions of the same page.\\
We want to identify the key UI differences between the two versions and then evaluate which variant is more effective UI/UX design that leads to better user experience and conversion.\\

You should\\
(1) Think step by step about the preliminary knowledge to answer the question, deconstruct the problem as completely as possible down to necessary sub-questions.\\
(2) With the aim of helping answer the original question, try to answer the sub-questions.\\
(3) Give your final answer according to the sub-questions and sub-answers.\\

You should end your answer with the following format (No bold, etc):\\
More effective: \textless{}First/Second\textgreater{}
\end{tcolorbox}
\caption{Prompt used for the UI/UX design selection task with DDCoT strategy.}
\label{prompt:ddcot_strategy}
\end{figure}

\begin{figure}[b]
\centering
\begin{tcolorbox}[
  colback=gray!5!white,
  colframe=gray!75!black,
  arc=2mm,
  fontupper=\small
]
You are an expert in designing UI/UX for web/apps. \\

And you are a debater. Hello and welcome to the debate.\\

The two screenshots show two different versions of the same page.\\
Identifying the key UI differences between the two versions, you think the \textbf{\{first/second\}} variant is more effective UI/UX design that leads to better user experience and conversion.\\
It's not necessary to fully agree with each other's perspectives, as our objective is to find the correct answer.\\

The opponent proposed that:\\  
\textbf{\{opponent\_opinion\}} \\
\end{tcolorbox}
\caption{Prompt used for the UI/UX design selection task with MAD strategy on the debate step arguing the first/second version better drives user actions. Prompt inputs are in \textbf{bold}.}
\label{prompt:mad_debater_first}
\end{figure}





\begin{figure}[b]
\centering
\begin{tcolorbox}[
  colback=gray!5!white,
  colframe=gray!75!black,
  arc=2mm,
  fontupper=\small
]
You are an expert in designing UI/UX for web/apps.\\

And you are a moderator. There will be two debaters involved in a debate.\\

The two screenshots show two different versions of the same page.\\
They are discussing which variant is more effective UI/UX design that leads to better user experience and conversion.\\
At the end of each debate round, you will evaluate answers and decide which is correct.\\

Now the \textbf{\{number\}} round of debate for both sides has ended.\\

Argue from the side arguing for the first version:\\
\textbf{\{first\_reason\}} \\

Argue from the side arguing for the second version:\\
\textbf{\{second\_reason\}} \\

Then what is the correct answer?\\
Please give the final answer that you think is correct and summarize your reasons.\\

You should end your answer with following the format (No bold, etc):\\
More effective: \textless{}First/Second\textgreater{}\\
\end{tcolorbox}
\caption{Prompt used for the UI/UX design selection task with MAD strategy on the moderation step. Prompt inputs are in \textbf{bold}.}
\label{prompt:mad_moderator_extractive}
\end{figure}

\begin{figure}[b]
\centering
\begin{tcolorbox}[
  colback=gray!5!white,
  colframe=gray!75!black,
  arc=2mm,
  fontupper=\small
]
You are an expert in designing UI/UX for web/apps.\\

The two screenshots show two different versions of the same page.\\
If the SECOND version has proven to be more effective UI/UX design that leads to better user experience and conversion, provide all possible and appropriate interpretations for its superiority.\\

\end{tcolorbox}
\caption{Prompt used for the UI/UX design interpretation task.}
\label{prompt:task2_zero_shot}
\end{figure}

\begin{figure}[b]
\centering
\begin{tcolorbox}[
  colback=gray!5!white,
  colframe=gray!75!black,
  arc=2mm,
  fontupper=\small
]
Given two statements are interpretations why one UI is better than another in terms of guiding user behavior effectively.\\

Statement 1:\\
\textbf{\{model\_reason\}}\\

Statement 2: \\
\textbf{\{reference\_reason\}} \\

Your task is to assess whether the first statement contains the same reasoning and essential content as the second statement. \\
If it does, answer with 'Yes'. Otherwise, answer with 'No'.\\

\end{tcolorbox}
\caption{Prompt used for GPT-4o-based evaluation on the UI/UX design interpretation task. Prompt inputs are in \textbf{bold}.}
\label{prompt:task2_gpt_eval}
\end{figure}












\clearpage

\clearpage